\documentclass{article}

\usepackage[preprint]{neurips_2021}

\setcitestyle{numbers,open={[},close={]}} %Citation-related 

\usepackage[utf8]{inputenc} % allow utf-8 input
\usepackage[T1]{fontenc}    % use 8-bit T1 fonts
\usepackage[pdftex]{hyperref}       % hyperlinks
\usepackage{url}            % simple URL typesetting
\usepackage{booktabs}       % professional-quality tables
\usepackage{amsfonts}       % blackboard math symbols
\usepackage{nicefrac}       % compact symbols for 1/2, etc.
\usepackage{microtype}      % microtypography
\usepackage{xcolor}         % colors

\usepackage{adjustbox}
\usepackage{wrapfig}
\usepackage{booktabs} % For formal tables
\usepackage{soul}
\usepackage{amsmath}
\usepackage{gensymb}
\usepackage{xfrac}
\usepackage{bm}
\usepackage{multirow}

\usepackage[shortcuts]{extdash} % import last!

\newcommand{\bv}[1]{\mathbf{#1}}
\newcommand{\doubleR}{\mathbb{R}}
\newcommand{\doubleZ}{\mathbb{Z}}
\newcommand{\doubleC}{\mathbb{C}}

\newcommand{\bra}{\langle}
\newcommand{\ket}{\rangle}

\newcommand{\mnistrot}{MNIST-rot}
\newcommand{\mnistrotlink}{\href{https://sites.google.com/a/lisa.iro.umontreal.ca/public_static_twiki/variations-on-the-mnist-digits}{\mnistrot}}
\newcommand{\modelnetforty}{ModelNet-40}
\newcommand{\modelnetfortylink}{\modelnetforty~\cite{ModelNet}}

\newcommand{\ReLU}{\operatorname{ReLU}}
\newcommand{\ELU}{\operatorname{ELU}}

\newcommand{\zseries}{{\mathrm{z}}}

\newcommand{\ok}[1]{#1}

\title{Nonlinearities in Steerable SO(2)-Equivariant CNNs}

\author{%
  Daniel Franzen \\
  Institute of Computer Science\\
  Johannes-Gutenberg University Mainz\\
  Staudingerweg 9,\\
   55122 Mainz, Germany\\
  \texttt{dfranz@uni-mainz.de}
  \And
  Michael Wand \\
  Institute of Computer Science\\
  Johannes-Gutenberg University Mainz\\
  Staudingerweg 9,\\
  55122 Mainz, Germany \\
  \texttt{wandm@uni-mainz.de}
}

\begin{document}

\maketitle

\begin{abstract}
	Invariance under symmetry is an important problem in machine learning. Our paper looks specifically at equivariant neural networks where transformations of inputs yield homomorphic transformations of outputs. Here, steerable CNNs have emerged as the standard solution. An inherent problem of steerable representations is that general nonlinear layers break equivariance, thus restricting architectural choices. Our paper applies harmonic distortion analysis to illuminate the effect of nonlinearities on Fourier representations of $SO(2)$. We develop a novel FFT-based algorithm for computing representations of non-linearly transformed activations while maintaining band-limitation. It yields exact equivariance for polynomial (approximations of) nonlinearities, as well as approximate solutions with tunable accuracy for general functions. We apply the approach to build a fully $E(3)$-equivariant network for  sampled 3D surface data. In experiments with 2D and 3D data, we obtain results that compare favorably to the state-of-the-art in terms of accuracy while permitting continuous symmetry and exact equivariance.
\end{abstract}

\section{Introduction}

Modeling of symmetry in data, i.e., the invariance of properties under classes of transformations, is a cornerstone of machine learning: Invariance of statistical properties over samples is the basis of any form of generalization, and the prior knowledge of additional symmetries can be leveraged for performance gains. Aside from data efficiency prospects, some applications require exact symmetry. For example, in computational physics, symmetry of potentials and force fields is directly linked to conservation laws, which are important for example for the stability of simulations.

In deep neural networks, (discrete) translational symmetry over space and/or time is exploited in many architectures and is the defining feature of convolutional neural networks (CNNs) and their successors. In most applications, we are typically interested in invariance (e.g., classification remains unchanged) or co-variance (e.g., predicted geometry is transformed along with the input). Formally, this goal is captured under the more general umbrella of equivariance~\cite{Cohen-GCNN-ICML-2016}:

Let $f: X \rightarrow Y$ be a function (e.g., a network layer) that maps  between vector spaces $X,Y$ (e.g., feature maps in a CNN). Let $G$ be a group and let (in slight abuse of notation) $g \circ v$ denote the application of the action of group element $g$ on a vector $v$. $f$ is called \textit{equivariant}, iff:
\begin{equation}
\forall g \in G: f(g \circ v) = h(g) \circ f(v),
\label{eq:equivariance}
\end{equation}
where $h: G \mapsto G'$ is a group homomorphism mapping into a suitable group $G'$. Informally speaking, the effect of a transformation on the input should have an effect on the output that has (at least) the same algebraic structure. Invariance ($h \equiv 1_{G'}$) and covariance ($h=id_{G\rightarrow G'}$) are special cases.%, along with contra-variance and any other isomorphisms of subgroups of $G$.

\textbf{G-CNNs:} To make current CNN architectures (consisting of linear layers and nonlinearities) equivariant, the standard (and also most general, see~\cite{Cohen-3DSteerable-2018}) approach are \textit{group convolutional networks} (G-CNNs)~\cite{Cohen-GCNN-ICML-2016}, which conceptually boil down to just applying all transformations $g \in G$ to filters, correlating the result with the data, and storing the results. Typically, $G$ will be a continuous, compact Lie group such as $SO(d)$ (our paper focuses on $SO(2)$). To avoid infinite costs, results are band-limited, stored as coefficients of a truncated Fourier basis on $G$. Simultaneously, the Fourier coefficients provide a linear representation of a subgroup of $G$ and thus exact equivariance in the sense of Eq.~\ref{eq:equivariance}. Using such a basis to directly construct sets of filters yields steerable filter banks~\cite{Cohen-Steerable-2017}, were each filter outputs a whole vector of coefficients that represent functions on $G$.

\textbf{Nonlinearities:} Unfortunately, band-limiting interferes in non-trivial ways with network architecture, as an application of standard nonlinearities such as $\ReLU, \tanh$, or even simple nonlinear polynomials to the Fourier coefficients will break equivariance.
Multiple solutions to this problem have been proposed~\cite{Weiler-generalE2-NeurIPS19}: Multiplicative non-linearities~\cite{Cohen-3DSteerable-2018,Kondor-covariantgraph-ICLR-2018} as in tensor networks keep equivariance but behave differently from traditional non-linearities and therefore cannot be used as a drop-in in classical CNN architectures. Complex nonlinearities such as $\doubleC$-$\ReLU$ that only act nonlinearly on the magnitude of Fourier coefficients \cite{worrall2017harmonic,Wiersma-SurfaceCNNs-Siggraph-2020} also keep perfect equivariance but are less expressive as they do not permit non-linear operations on the phase information. A recent study \citet{Weiler-generalE2-NeurIPS19} shows that simple discretized rotations \cite{Cohen-GCNN-ICML-2016}, which do not require architectural adaptations but provide only approximate equivariance, yield the best practical results in image classification tasks.	

\textbf{SO(2)-equivariance:} The goal of our paper is to clarify the effect of non-linearities on Fourier-domain representations. We restrict ourselfs to the case of $SO(2)$, which permits the application of standard harmonic distortion analysis~\cite{RingingReLUs-AliMehmetiGoepel-2021}. Our goal is to maintain a band-limited representation of a function on $SO(2)$ (corresponding to a fixed angular resolution) and efficiently compute the a band-limited Fourier-representation after application of a non-linearity. We obtain an exact algorithm for polynomial nonlinearities with a computational overhead of $\mathcal{O}(D \log D)$ for degree $D$. For general nonlinearities, we can formally motivate a \ok{quick} convergence, which we validate in numerical experiments. The effect of this overhead in practice is usually minor.

\textbf{SE(3)-equivariant surface networks:} While the limitation to $SO(2)$ might appear restrictive, it is still important for many problems in processing image and geometric data: Adding translational invariance ($E(2)$-equivariance) is easy, and we also apply our representation to surface data with normal information, extending ideas of Wiersma et al.~\cite{Wiersma-SurfaceCNNs-Siggraph-2020} to a fully $E(3)$-equivariant network

We evaluate our networks on example benchmarks for 2D image and 3D object recognition. We obtain invariant results and equivariant intermediate representations up to numerical precision for polynomial nonlinearities (double checking the formal guarantees) and low-error approximations in with general nonlinearities such as $\tanh$,$\ReLU$,$\ELU$~\cite{clevert2016fast} at reasonable overhead (less than 20\%, depending on approximation quality). Classification accuracy on {\mnistrotlink} and {\modelnetfortylink} is on par with the state-of-the-art for $\ReLU$ and slightly reduced for low-degree polynomials~\cite{Gottemukkula2020PolynomialActications}.

Our main contributions are (i) a simple analytical model for the effect of non-linearities on Fourier representations in $SO(2)$-equivariant networks and (ii) an efficient algorithm for applying nonlinearities. It is provably exact for polynomials and empirically yields good approximations for common non-polynomial functions. This permits, for the first time, a the usage of standard CNN architectures with common nonlinearities without compromising equivariance.

% Head 1
\section{Related Work}
\label{sec:related-work}

\textbf{Equivariant networks:} The are various approaches to achieving equivariance or invariance to rotations and other transformations in CNNs. The first approaches focused on exact $C_4$-equivariance of rotated images~\cite{laptev2016tipooling} by applying the same network to rotated copies of the input (with shared weights), followed by some invariant operation (e.g. max-pooling), after which the feature maps become invariant to input rotations. More advanced architectures allow the network to ``see'' filter responses of the previous layer to different rotations of the inputs by adding weights for those inputs in a way that does not break equivariance \cite{Cohen-GCNN-ICML-2016,Cohen-Steerable-2017,dieleman2016exploiting}. More fine-grained equivariance can be achieved by representing the image on a hexagonal grid~\cite{hoogeboom2018hexaconv}, or by using \emph{steerable filters}~\cite{weiler2018learning} and Fourier representations~\cite{worrall2017harmonic}. A comprehensive study that puts these different approaches into a common framework is provided by \citet{Weiler-generalE2-NeurIPS19}.

\textbf{3D data:} While early neural network models for 3D data operated on regular grids similar to their 2D counterparts, alternative methods for processing 3D data with neural networks have quickly emerged to avoid the overhead. Some methods rely on data reduction through lower dimensional projections, such as generating rendered images (sometimes from multiple perspectives to improve performance and approximate equivariance) and feeding them into into classical 2D-CNNs~\cite{Su2015MultiviewCN}. Other methods employ projections on spherical surfaces, on which convolution operations can be defined. These surfaces can be represented by a sampling (e.g. CSGNN\cite{CSGNN} uses an icosahedral sampling) or a spherical harmonics basis~\cite{SphericalCNNs}.

Graph-based networks perform convolutions on a connectivity graph. For example, SchNet~\cite{SchNet} is a popular architecture for predicting molecular properties, operating on the graph of molecular bonds. Another example is MeshCNN~\cite{MeshCNN}, which defines convolution operations on vertices of a 3D triangle mesh and uses a vertex merging as pooling operation. Other methods work directly on the intrinsic geometry of manifolds\cite{Wiersma-SurfaceCNNs-Siggraph-2020}. Such methods often have the advantage of being naturally equivarient to rotations or translations, as they work on intrinsic properties of an object. However, they often need to be tailored to a specific type of data.

Point based architectures have become a popular alternative. Early models like PointNet~\cite{PointNet} and follow-ups~\cite{PointNetPP,PointCNN} directly used the coordinates of 3D data as network input. Later, Point Convolutional Neural Networks~\cite{PCNN} emerged as an attempt to generalize the grid-based CNNs architecture to a spatial convolution on point clouds, using a radial Gaussian filter basis with fixed or trainable offset values. Kernel Point Convolutions (KPConv~\cite{KPConv}) improve on this by introducing correlation functions which define the interaction of nearby points and limiting the distance of interactions to reduce overhead.

There are various ways of transforming these architectures to rotationally invariant models. One way is use layers that produce rotationally invariant features, which can then be processed further without restrictions, for example by aligning the inputs to the convolutional filters. This approach is taken by GCANet~\cite{GCANet}, and by MA-KPConv~\cite{MAKPConv}, which extends the KPConv model by using multiple alignments for the filters. Another approach is to apply an \emph{invariant map} as the final function of each layer, as in Spherical Harmonics Networks (SPHnet)~\cite{SPHnet}, which calculate activations in a spherical harmonic basis and produce invariant output by taking the norm over coefficients with identical degree. Various papers generalize the notion of steerable filters to 3-dimensional data~\cite{Cohen-3DSteerable-2018,TensorFieldNetworks}.

Our $SE(3)$-architecture is probably closest to the method of \citet{Wiersma-SurfaceCNNs-Siggraph-2020}, which differs from the methods named above by giving each feature vector its own local reference frame, which is aligned to the surface normal and one arbitrarily chosen perperdicular direction. Equivariance is guaranteed by using parallel transport along the surface to align the reference frames of different features. In contrast to this, we skip the parallel transport step and simply rotate the normal vectors onto each other to find a suitable alignment of the local coordinate systems.

Our main analytical tool is harmonic distortion analysis. Originally developed in physics and engineering~\cite{feynman-lectures-1965}, Ali Mehmeti-Göpel et al.~\cite{RingingReLUs-AliMehmetiGoepel-2021} have recently applied it to the problem of understanding the trainability of deep networks. In our case, we study how our linear representations of $SO(2)$ are affected by using a similar transformation with respect to an angular variable $\alpha$.

\section{SO(2)-Equivariant Networks}
\label{sec:so2-equivariance}

We start with a very brief recap of $SO(2)$-equivariant steerable networks \cite{Cohen-Steerable-2017}. We formulate the approach in terms of band-limited angular functions rather than steerable filter banks. This is merely a transposed view but facilitates the discussion in Section~\ref{sec:nonlinearities}.

%\subsection{Mapping Functions on SO(2)}

The starting point is a network layer $f^{(l)}, l\in\{1...L\}$ that maps functions $x^{(l-1)}$ to functions $x^{(l)}$, both having the domain $SO(2)$ and vector-valued output. We use an arc-length parametrization
\begin{equation}x^{(l)}:\doubleR \rightarrow \doubleR^{d_{l}}\text{ \ \ with periodicity \ \ } x^{(l)}(\alpha+2\pi) = x^{(l)}(\alpha) \text{ for all }\alpha \in \doubleR\end{equation}
To make layer activation functions representable in finite memory, we assume that each is band-limited to a maximum frequency of $K^{(l)}$ and thus can be represented by a complex Fourier series
\begin{equation}
x^{(l)}(\alpha) = \sum_{k=-K_l}^{K_l} \bv{z}_k^{(l)} \cdot e^{ik\alpha},\qquad \bv{z}_k^{(l)} \in \doubleC^{d_l},\ \  \bv{z}_k^{(l)} = \overline{\bv{z}}_{-k}^{(l)}.
\end{equation}
The conjugation symmetry holds because we are representing real functions. In a concrete implementation, we therefore store only half of the coefficients. According to the sampling theorem~\cite{Glassner-ImageSynthesis-1995}, such functions can be represented exactly by a uniform sampling with $2K_l+1$ samples, and the mapping between the two discrete representations of $2K_l+1$ coefficients each is a unitary bijection.
The layer function $f^{(l)}$ is, as usual, a concatenation of a linear function $W^{(l)}$ and a point-wise non-linearity $\varphi:\doubleR \rightarrow \doubleR$ that is applied to all angles and all output dimensions:
\begin{equation}
	\big[[f^{(l)}(x^{(l-1)})](\alpha)\big]_i = \varphi\big(\big[[W^{(l)} (x^{(l-1)})](\alpha)\big]_i\big)
	\label{eq:pointwise-nl}
\end{equation}
Above, we use ``$[\ \cdot\ ]_i$'' to denote indexing of vector-valued function outputs. 

%\paragraph*{Linear equivariance:}
We now construct linear $W^{(l)}$ that are equivairant under rotations. In our representation, these are cyclic shifts, i.e., equivariance can be expressed as $W(x(\alpha+T))=(W\circ x)(\alpha+T))$ for all $T\in\doubleR$. This means that $W$ is a shift-invariant linear operator. Signal theory~\cite{Glassner-ImageSynthesis-1995}  tells us that these correspond do convolution of the input function with a kernel $w_i^{(l)}:[0,2\pi]\rightarrow \doubleR^{d_{l-1}}$. A subtlety: Considering input angles $\alpha$ and output angles $\beta$, we can introduce a dependency of $w_i^{(l)} = w_{i,\beta}^{(l)}$ on the output parameter while maintaining equivariance wrt.~$\alpha$. \citet{Weiler-generalE2-NeurIPS19} show that this is not only the most case and but also provides non-trivial performance benefits in practice:
\begin{equation}
	[W^{(l)} (x^{(l-1)})]_i(\beta) := \int_{0}^{2\pi} \bra x^{(l-1)}(\alpha), w_{i,\beta}^{(l)}(\alpha-\beta) \ket d\alpha 
	\label{eq:simple-convolution}
\end{equation}
Due to the convolution theorem, this can be more conveniently written as a point-wise multiplication of the output Fourier coefficients $\bv{z}_k^{(l-1)}$ with the coefficients $\bv{q}_{k,k',i,j}^{(l)}$ of the a Fourier-series of $w_i^{(l)}$
\begin{equation}
	z_{k',i}^{(l)} = \sum_{k=1}^{K_{l-1}} \sum_{j=1}^{d_{l-1}} q_{k, k',i,j}^{(l)} \cdot z_{k,j}^{(l-1)}.
\end{equation}
The index $k$ refers to the input and $k'$ to the output frequency, and $i$ and $j$ the input/output feature indices. We use the $K_l \times K_{l-1} \times d_l \times d_{l-1}$-tensor $\bv{q}^{(l)}$ directly as trainable parameters for each layer.
Overall, the application of a $W$ to an $x$ is structurally simple, requiring only a (complex) linear map of the input Fourier coefficients. Issues, however, arise when trying to apply the nonlinearity $\varphi$.

\section{Nonlinearities in $SO(2)$-Equivariant Networks}
\label{sec:nonlinearities}

In the continuous case, Eq.~\ref{eq:simple-convolution} settles the problem: 
Equivariance is still holds (trivially) if the non-linearity $\varphi$ is applied point-wise, at every angle, as defined in Eq.~\ref{eq:pointwise-nl}. The issue is that this requires not only a transformation back from the frequency into the angular domain, but also a re-encoding, which involves a continuous integral that requires a potentially expensive numerical calculation: Unlike linear mappings, a non-linear can create higher-order frequencies, called ``harmonics''.

\subsection{Non-Linearities Create Harmonics}

In order to understand this effect, we apply harmonic distortion analysis \cite{feynman-lectures-1965,RingingReLUs-AliMehmetiGoepel-2021}.
To simplify the notation, we will in the following drop the layer and feature index and denote by $z_k, k\in\{-K,...K\}$ the Fourier coefficients of the \textit{preactivation} $[W^{(l)}(x^{(l)})]_i$ of a single, 1D feature channel at a fixed layer. We denote the whole series as $\zseries=[z_{-K},...,z_K]$. For the analysis, we also start by assuming that $\varphi$ is a polynomial of finite degree $D$:
\begin{equation}
	\varphi(x) = \sum_{j=0}^D t^j x^j
\end{equation}
Plugging in the corresponding Fourier series yields:
\begin{eqnarray}
f(\alpha) &=& \varphi\left(\sum_{k=-K}^K z_{k} e^{ik\alpha}\right)  %\\
%&=&%\scriptstyle
=
\sum_{j=1}^D \ t_j \cdot \sum_{k_1,...,k_j=-K}^K z_{k_1} \cdots z_{k_j} e^{i(k_1 + \cdots + k_j)\alpha}, 
\label{eq:large-poly}
\end{eqnarray}
The convolution theorem now converts the point-wise multiplication into a convolution of the spectra in the Fourier domain: The series $\zseries' \in \doubleC^\doubleZ$ of output Fourier coefficients is given by
\begin{equation}
\zseries' = t_0 \ + \  t_1 \zseries \ + \  t_2 (\zseries \otimes \zseries) \ + \  t_3 (\zseries \otimes \zseries \otimes \zseries) \ + \  \cdots \ + \  t_D (\zseries \otimes \cdots \otimes \zseries)
\label{eq:convolve}
\end{equation}
where ``$\otimes$'' denotes the discrete convolution 
\begin{equation}
[\mathrm{z} \otimes \mathrm{w}]_k := \sum_{m\in\doubleZ} z_{m} \cdot w_{k-m}.
\label{eq:def-discrete-conv}
\end{equation}
As expected, an application of a non-linear function could potentially spread the spectrum towards higher frequencies. Inputs band-limited to frequency $K$ yield outputs will band-limited to $KD$. 

We can use this observation to construct efficient algorithms for computing $\zseries'$ from $\zseries$.
%
%\subsection{Efficient Application of Non-Linearities}
%
Our goal is to compute the first $K_l$ Fourier coefficients efficiently from the $K_{l-1}$ Fourier coefficients of layer $l-1$. The linear layer corresponds to a simple complex matrix-vector multiplication. However, applying the activation function is not trivial, due to the presence of negative frequencies in the Fourier series, that lead to a mixing of high and low frequency components. Further, a naive computation on only $K$ coefficients would introduce aliasing effects that break equivariance. 

\subsection{Exact Equivariance: Polynomial NonLinearities}
\label{sec:polyfft}
%\subsection{Applying Nonlinearities}

\textbf{Direct Convolution:} The simplest correct solution is to directly and iteratively evaluate the discrete convolution in Eq.~\ref{eq:convolve} $D$-times,
requiring $\mathcal{O}(\sum_{j=1}^D K_{l-1} \cdot j  K_{l-1})  %= \mathcal{O}(K^2\sum_{j=1}^D j )
= \mathcal{O}(K_{l-1}^2 D^2)$ time and $\mathcal{O}(DK_{l-1})$ temporary memory, with only $\mathcal{O}(K_{l})$ values being kept in the end.

\textbf{FFT-Based Algorithm:} The direct algorithm becomes inefficient for higher-order polynomials. The alternative is to evaluate $\varphi(f(\alpha))$ directly in the angluar domain, which requires an inverse Fast Fourier Transformation (IFFT), application of $\varphi$ and an forward FFT. For large $D$, the point-wise application of $\varphi$ in the anbular domain is obviously more efficient than the Fourier-domain convolution. However, in order to maintain equivariance, we need to make sure to sample adequately.

Counting the non-zero coefficients in Eq.~\ref{eq:def-discrete-conv} shows us immediately that $2DK_{l-1}+1$ Fourier coefficients, i.e., a Fourier expansion up to frequency $DK_{l-1}$, is sufficient for an exact evaluation of a degree $D$ polynomial $\varphi$: Arising from a $D$-fold convolution in the Fourier domain, the signal is band-limited accordingly. The sampling theorem then translates this to equidistant sampling at $2DK_{l-1}+1$ discretization points in the angular domain (the corresponding discrete FFT becomes a bijection). This gives us an asymptotically more efficient algorithm with run-time $\mathcal{O}(DK_g \log DK_g)$ and memory $\mathcal{O}(DK_g)$ that is still exact. As the discrete Fourier transformation is also a unitary map, we can also expect favorable numerical properties. For low orders $D$, direct covolution might be slightly more efficient than the FFT-based approach. Nonetheless, our current implementation uses only the FFT variant, for simplicity, and also because we observe that, anyways, the overhead of applying the non-linearities is minor in the overall computational costs in our networks.

\textbf{Practical application:} In practice, we use polynomial non-linearities of moderate degree, as high-order polynomials become unstable~\cite{Gottemukkula2020PolynomialActications} (computational costs of increasing $D$ are not a limiting factor). A problem of is that any polynomial of degree $D\geq 1$ diverges asymptotically, thus making training unstable. We address this by clip the $\ell_1$-norm of the Fourier coefficients at a maximum value, keeping $||z||_1\leq c$ with $c$ chosen bounding the range the polynomial was designed for. The $\ell_1$-norm is an upper bound of the maximum value $x(\alpha)$ for all $\alpha$ that is also tight, as can be seen by an example Fourier series with aligned complex phases such that $|x(0)|=\sum_{k=-K}^K |z_k|$.

\subsection{Approximation: Oversampled FFT}

For non-polynomial nonlinearities $\varphi$, the FFT-based algorithm can still be used for approximate evaluation. We still map to the angular domain and back via FFT, using $D$-fold oversampling, but $D$ now becomes a user-parameter. In the general case, the spectrum will not be band-limited, but decaying (the Fourier series converges for functions of bounded variation, which certainly applies to post-activations in all practical networks). Thus the truncated FFT will create (non-equivariant) aliasing artifacts that should vanish with increasing $D$.

The convergence behavior depends on the fall-off of the Fourier spectrum for a non-linearity that is not bounded in polynomial degree, which is hard to quantify. Even if we assume a polynomial approximation, we would need large degrees and results would depend on the decay rate of the $t_j$ employed to gain a realistically tight bounds. Empirically, we do observe an exponential increase of precision with oversampling, as shown experimentally in Section~\ref{sec:results}. As the run-time costs of the FFT are moderate, this still a practical algorithm.

\section{Application to Geometric Data}

In the following, we apply the $SO(2)$-equivariant network layer to 2D and 3D data.

\begin{figure}[t]
	\adjustbox{valign=t}{\begin{minipage}[b]{0.45\textwidth}
			\begin{tabular}{cc}
				\raisebox{4mm}{\includegraphics[scale=0.15]{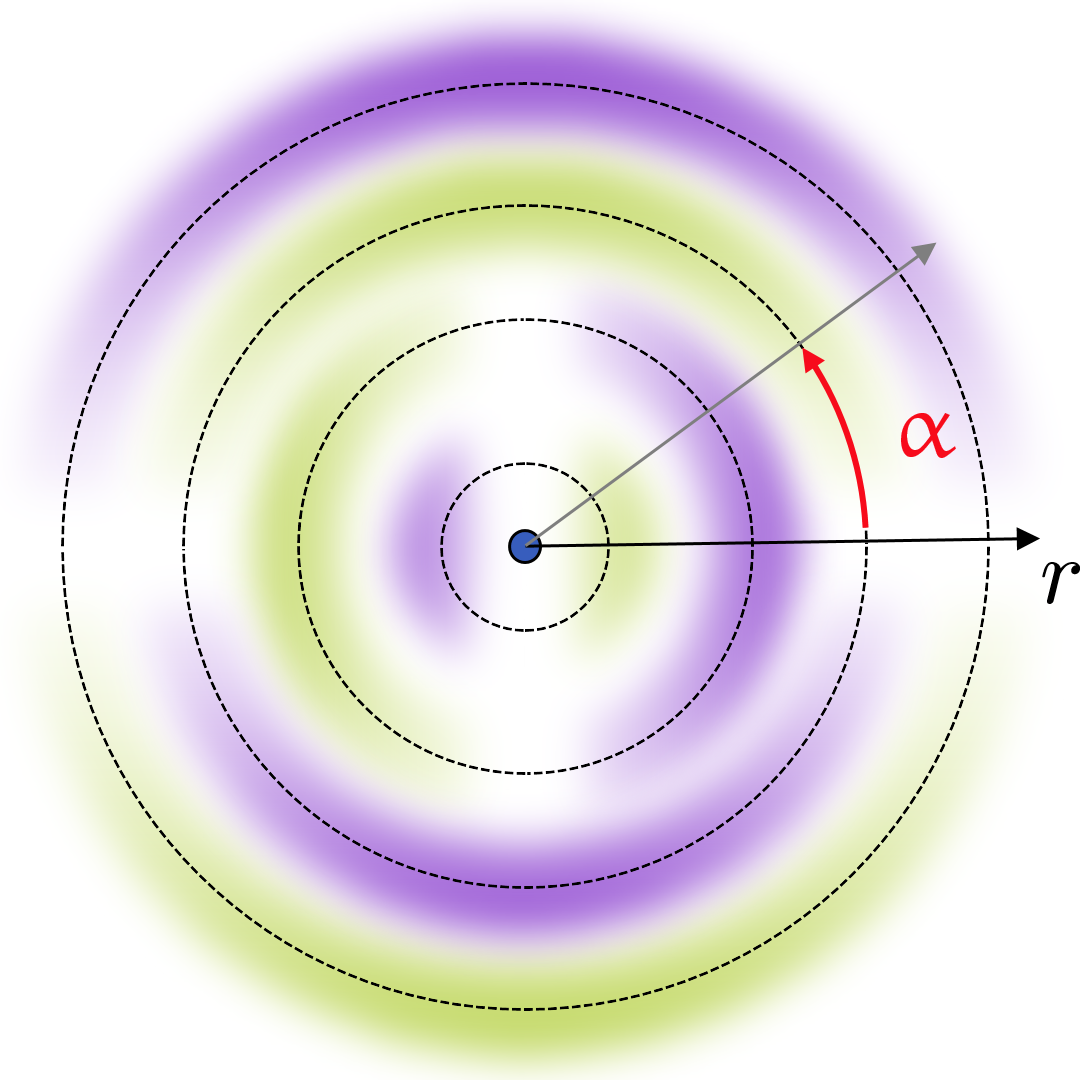}}
				&
				\includegraphics[scale=0.18]{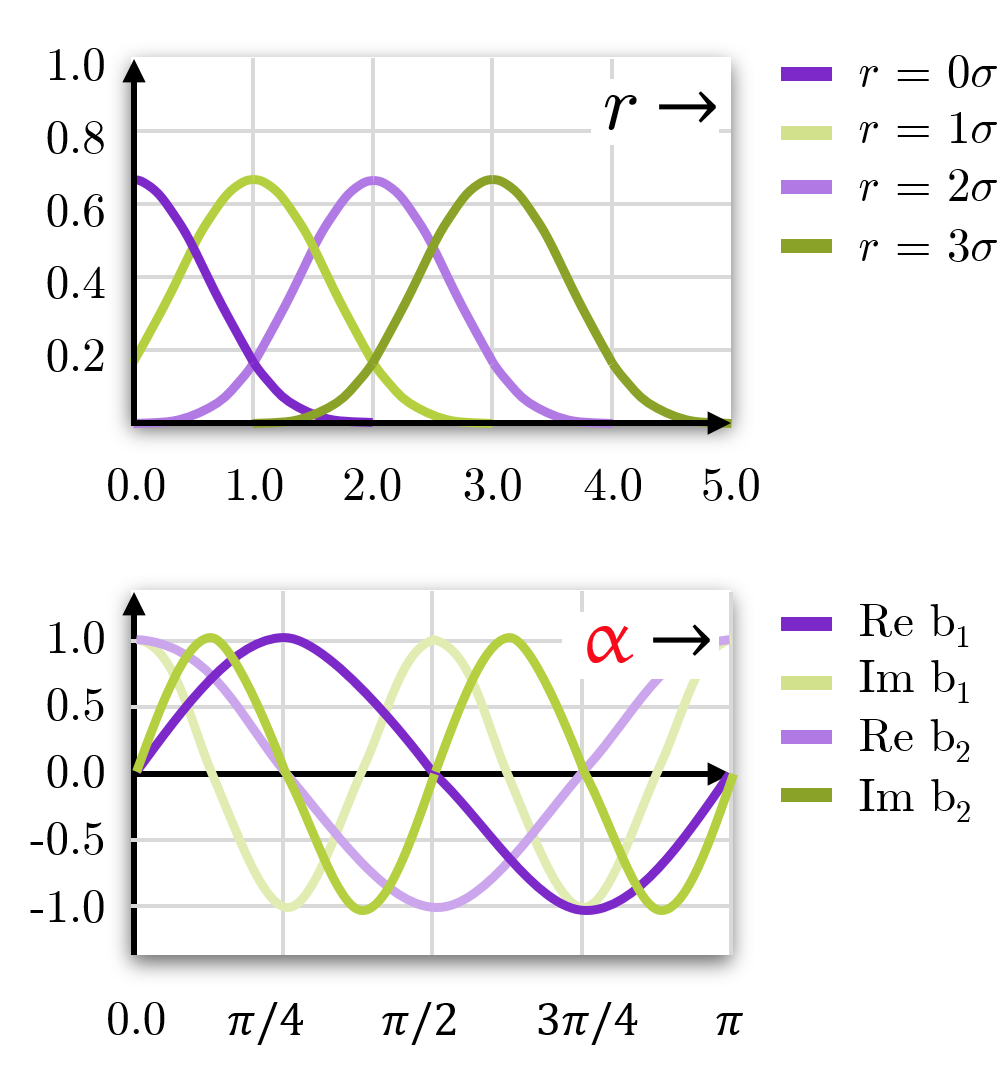}
			\end{tabular}
			\caption{\textit{$SE(2)$-equivariant point cloud networks:} Each point is associated with a set of concentric $SO(2)$-steerable filters (Fourier basis), modulated by equidistant gaussians in radial direction (for translational band-limiting).}
			\label{fig:basis}
	\end{minipage}} \hfill %
	\adjustbox{valign=t}{\begin{minipage}[b]{0.463\textwidth}
			\includegraphics[width=\columnwidth]{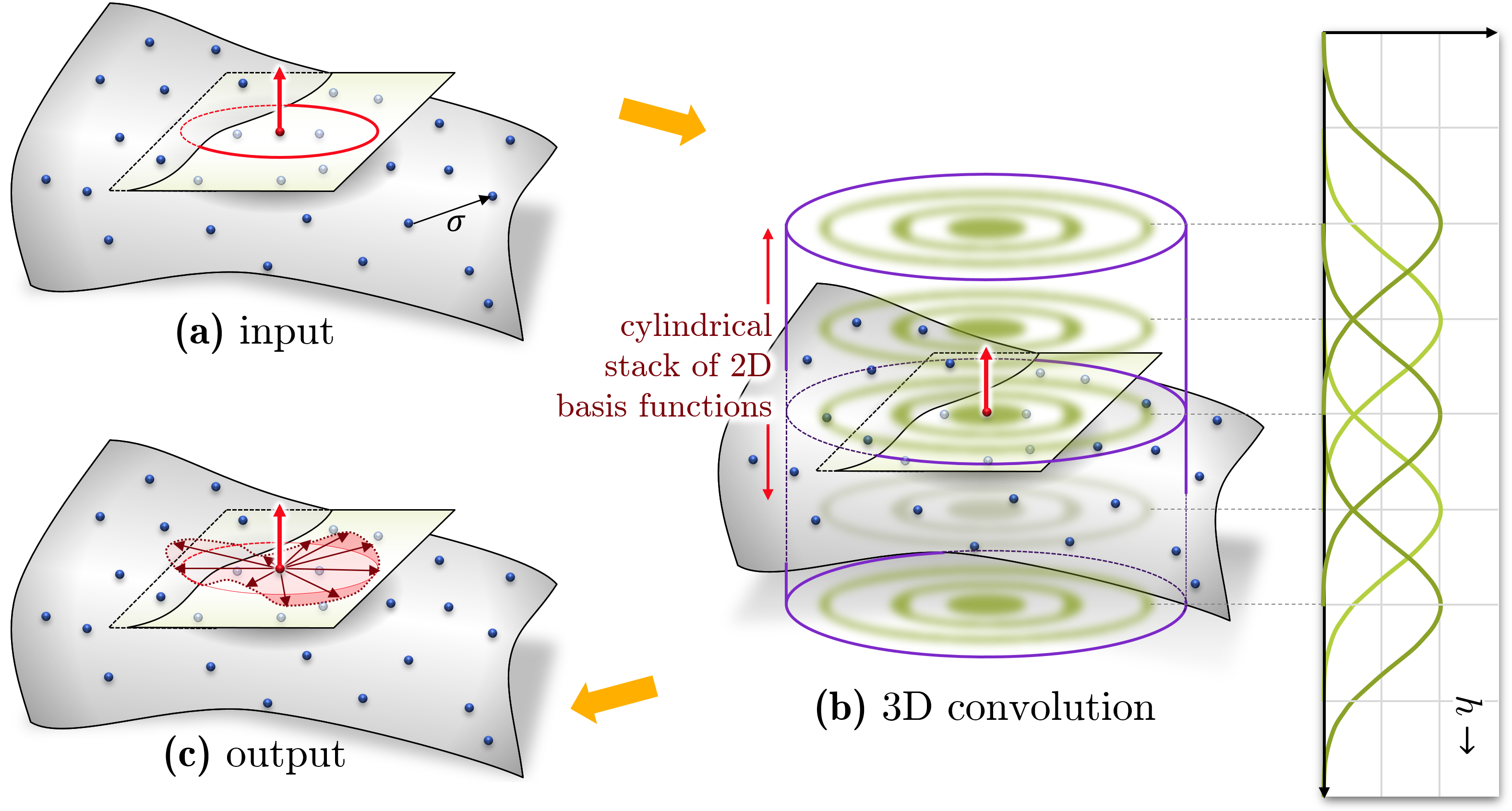}%& 
			\caption{\textit{$SE(3)$-equivariant surfel networks}: (a) for oriented surfels, we (b) perform 3D convolutions with unknown rotation around the normal direction, outputing (c) a collection of functions that assign scalar to tangent vectors.}
			\label{fig:surfel-network}
	\end{minipage}}
\end{figure}

\subsection{SE(2)-Equivariant Networks for 2D Point Clouds}

We construct an SE(2)-equivariant network for a set of input points $\bv{p}_1,...,\bv{p}_n$ that carry arbitrary attribute vectors $\bv{y}_i$. These points can be a regular pixel grid of an image or a sampling of a geometric object. We equip each point $\bv{p}_i$ with a set of radial filters, in polar coordinates $(r,\varphi)$ such that a 2D filter corresponds to a set of concentric circles of radius $r>0$, on each of which a Fourier basis in $\varphi$ is used to linearly represent $SO(2)$ (Fig.~\ref{fig:basis}). Radially, we discretize by simple equidistant FIR low-pass filters --- our current implementation uses equidistant Gaussians 
$\omega_m(r) := \exp(-(\frac{r}{2\sigma}-m)^2)$:
\begin{equation}
\label{eq:filter-basis}
b^{(l)}_{k,m}(\varphi,r) := \omega_m(r) \cdot e^{2ik\varphi}, \ \ m =0,1,...,M^{(l)}, \ k = -r^{(l)} \cdot m,...,r^{(l)} \cdot m
\end{equation}
The whole construction is illustrated in Figure~\ref{fig:basis}. The input points for each layer are treated as Dirac functions $x^{(0)}(\bv{t}) = \sum_{i=1}^n \bv{y}_i\delta_{\bv{p}_i}( \bv{t} )$.

\textbf{Further layer types:} To build complete networks, we employ spatial average pooling layers, which averages Fourier coefficients (which is possible due to linearity of the Fourier transform). For the projection to invariant features, we either output only the scalar Fourier coefficient $z_0$ in the last convolutional layer (\emph{conv2triv}) or take the norm of all complex outputs of the last convolutional layer after the nonlinearity has been applied.\\
Batch-normalization~\cite{Batchnorm-Ioffe-ICML-2015} of Fourier-representations follows the obvious route of obtaining the mean via the $z_0$-coefficients and the variances via the power spectrum $||\zseries||_2^2$ of the Fourier coefficients. Instead of tracking a running mean during training for batch normalization, we calculate the exact training set statistics after training is done in one extra pass, while not changing other network weights.

\begin{figure}[t]
	\adjustbox{valign=t}{\begin{minipage}[b]{0.5\textwidth}
			\begin{centering}
				\vspace{2mm}
				\begin{tabular}{cc}
					\includegraphics[width=0.4\linewidth]{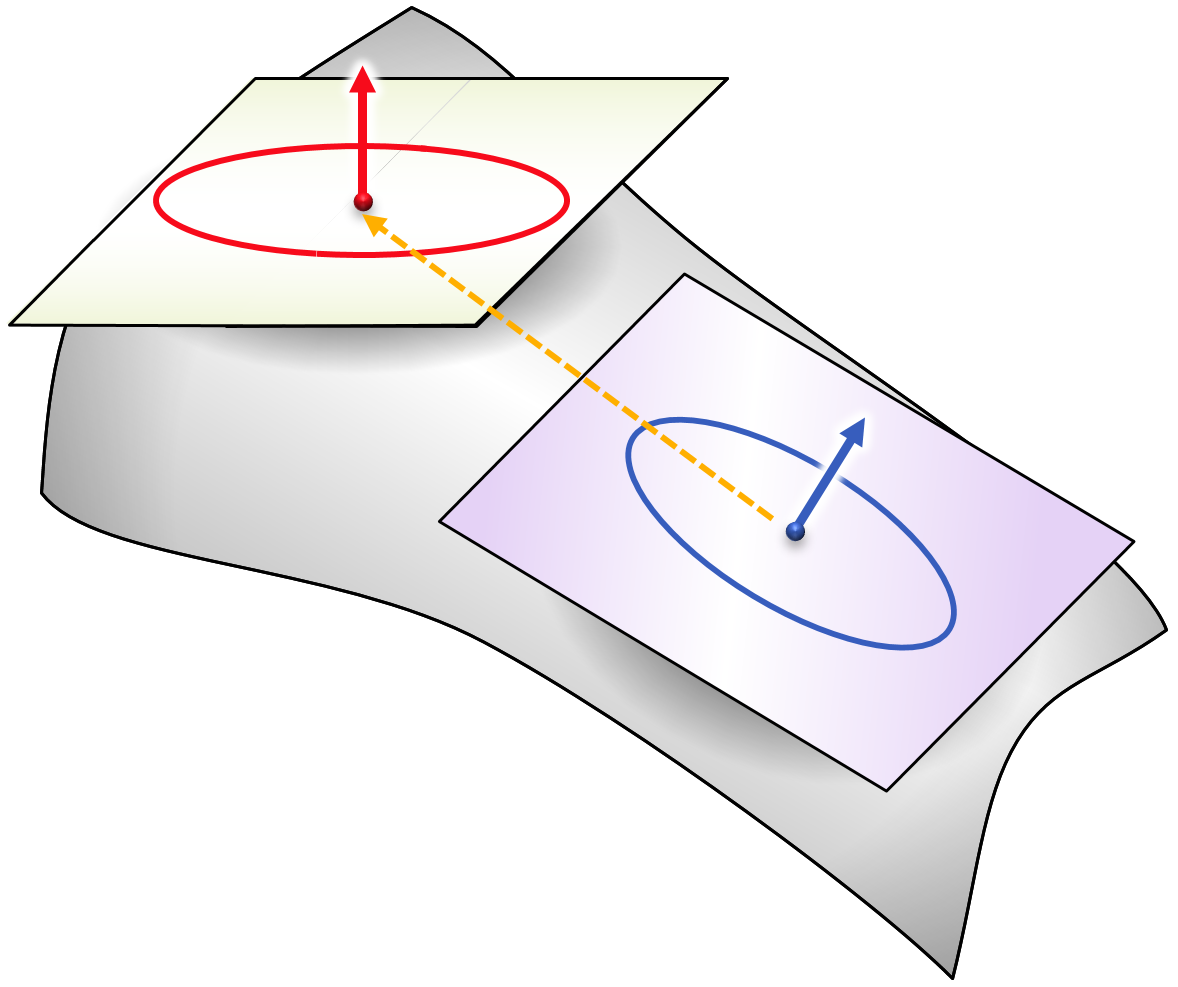} &
					\includegraphics[width=0.5\linewidth]{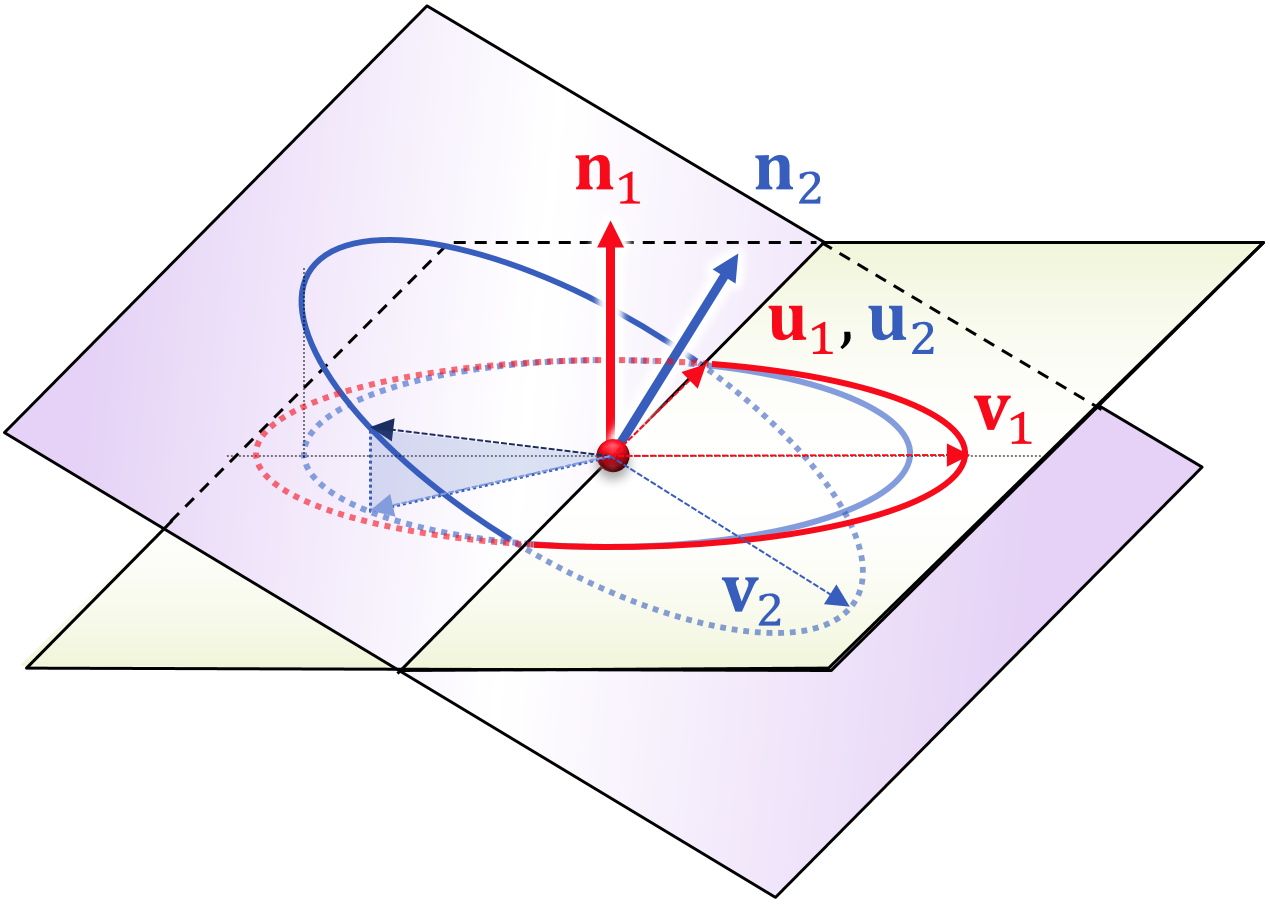}
				\end{tabular}
				\vspace{3mm}\caption{The output feature functions, which are tangent vectors, are aligned by projecting into the tangent-space of the target point.}
				\label{fig:proj-tangent}
			\end{centering}
	\end{minipage}} \hfill %
	\adjustbox{valign=t}{\begin{minipage}[b]{0.4\textwidth}
			\begin{centering}
				\includegraphics[width=0.94\linewidth]{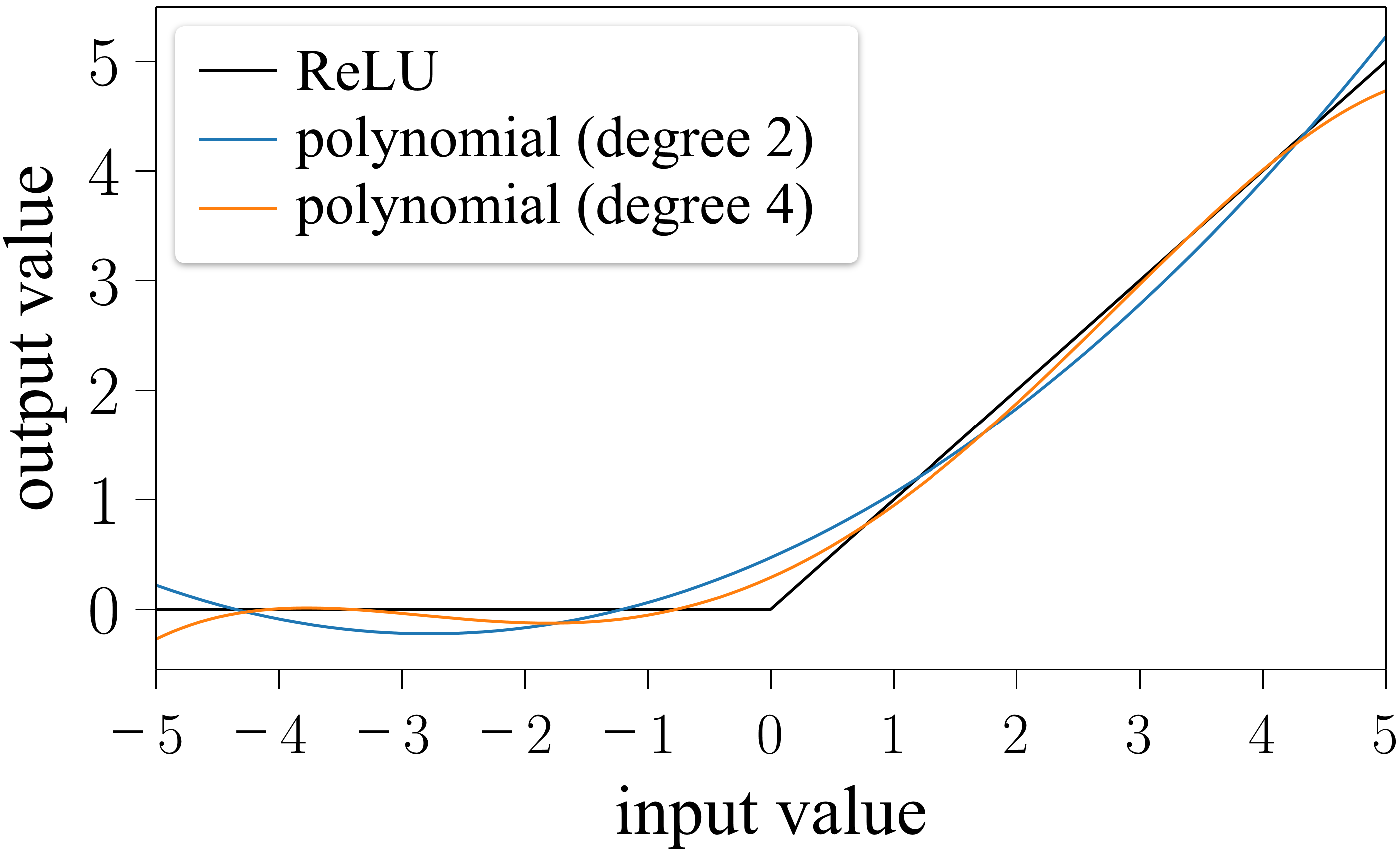}
				\caption{Polynomial approximations of the ReLU function~\cite{Gottemukkula2020PolynomialActications}.}
				\label{fig:polyReLUapprox}
			\end{centering}
	\end{minipage}} \hfill %
\end{figure}

\textbf{Non-linearities:} We evaluate our network for various general nonlinearities ($\ReLU$, $\tanh$ and $\ELU$~\cite{clevert2016fast}), as well as polynomial approximation of the ReLU function of degrees 2 and 4 (see Figure~\ref{fig:polyReLUapprox}) taken from \citet{Gottemukkula2020PolynomialActications}, computed by FFT algorithm outlined in section~\ref{sec:nonlinearities}. In case of the polynomial activations, we clamp the L1-norm of each channel's Fourier coefficients to the range $[-5, 5]$ before applying the nonlinearity to avoid problems with exploding activations or gradients. We also include the $\doubleC$-$\ReLU$ function in our experiments (which acts on the norm of the activations only and therefore requires no Fourier transformation) to better estimate the performance penalties of the FFT method.

\textbf{Architecture:} Our concrete network design follows construction of \citet{Weiler-generalE2-NeurIPS19} for their best {\mnistrot}-models, we use the same number of equivariant and linear layers with the same output channel count and filter properties (radii, rotation orders and width) and also apply Dropout~\cite{Srivastava2014DropoutAS} with $p=0.3$ before each linear layer. We train our models for 40 epochs with at batch size of 64 images. We use the Adam~\cite{Kingma2015AdamAM} optimizer, starting with a learning rate of 0.015, with an exponential decay factor of 0.8 per epoch starting after epoch 16. We calculate the mean test error and its standard deviation from 10 independent training runs with He-initialized~\cite{HeInit} weights. 

\subsection{$SE(3)$-Equivariant Surfel Networks}
\label{sec:SurfelNetworks}

Following the concept of \citet{Wiersma-SurfaceCNNs-Siggraph-2020}, we can apply $SO(2)$-equivariant layers for building $SE(3)$-equivariant surface networks (our construction differs slightly: it uses extrinsic rather than intrinsic operations, thus actually creating exact $SE(3)$-equivariance but not having isometric (``bending'') invariance). The main idea is that we only consider point-sampled surfaces with oriented normals at every point (``surfels'', \cite{Pfister-Surfels-2000}). We then perform the same construction as in the 2D point cloud case, equipping each individual point with vertical stack of radial filters in its tangent plane (see Fig.~\ref{fig:surfel-network}a,b). In vertical direction, we use the same Gaussian filters as in radial direction (for consistent band-limiting). The filters are convolved against the point clouds, with uniform features $\bv{y}_i=1$ for all $i$. The resulting angular functions can be interpreted as a vector-valued function in the tangent plane (Fig.~\ref{fig:surfel-network}c).

When performing a second level of convolution on already complex-valued input coefficients, we need to relate these angular feature functions computed by different surfels, which live in different reference frames. Here, we project the tangent vectors into the tangent plane, see Fig.~\ref{fig:proj-tangent}. This construction is purely geometric and thus covariant under rigid transformations (the whole geometry is just rotated/translated together). Note: Aiming at different applications, \citet{Wiersma-SurfaceCNNs-Siggraph-2020} use intrinsic parallel transport here, which is more invariant but requires manifold input.

\section{Results}
\label{sec:results}

\begin{figure}[t]
	\adjustbox{valign=t}{\begin{minipage}[b]{0.48\textwidth}
			\includegraphics[height=5.4cm]{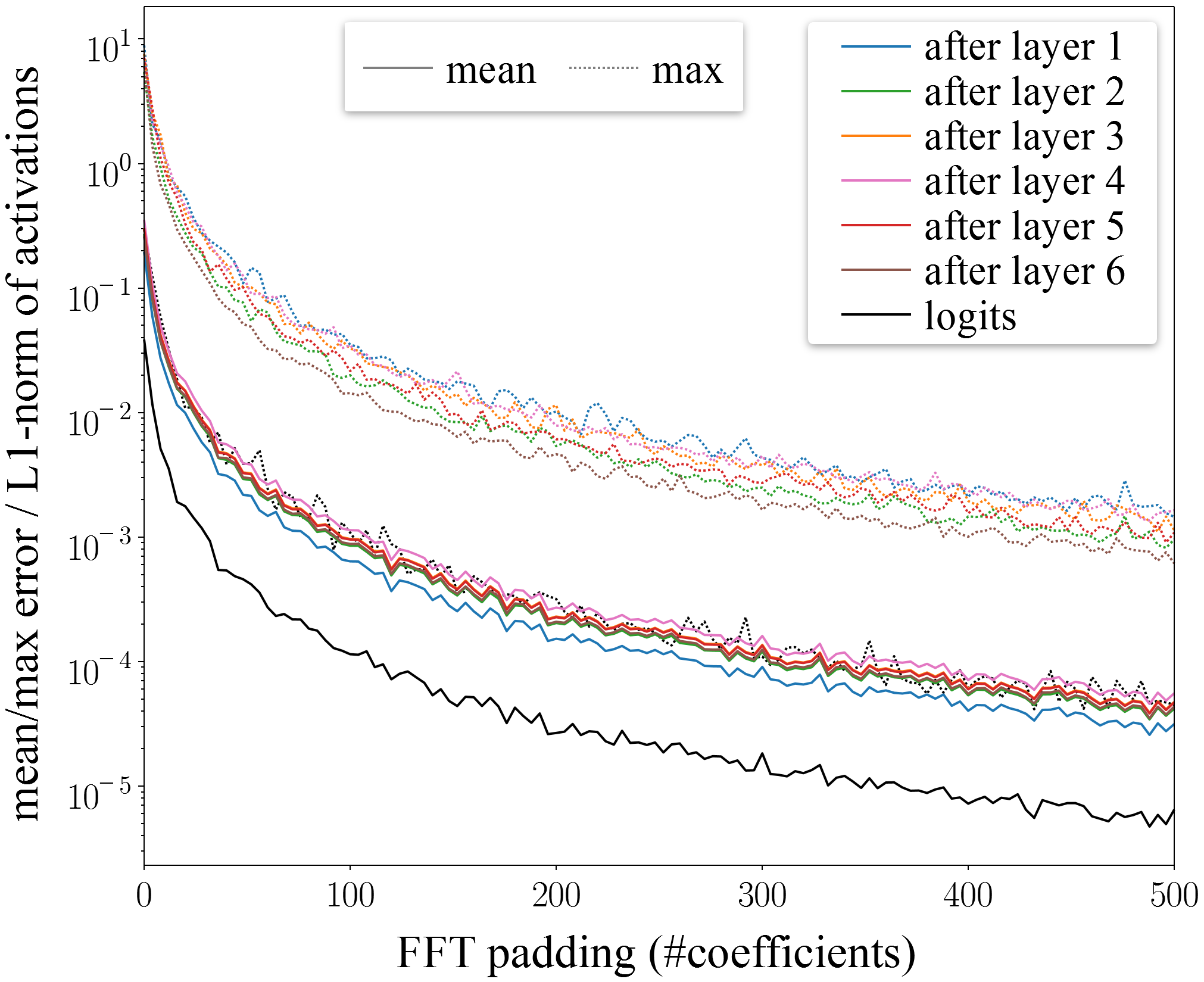}
			\caption{Relative error of  ReLU activations (basic {\mnistrot} network, 9 Fourier coeffs., \emph{norm}-map) for random rotations vs. unrotated input.
			solid: mean absolute error, dashed: maximum error, relative to the layer-wise L1-norm for batches of 32 images.}
			\label{fig:errorReLU}
	\end{minipage}} \hfill %
	\adjustbox{valign=t}{\begin{minipage}[b]{0.48\textwidth}
			\includegraphics[height=5.4cm]{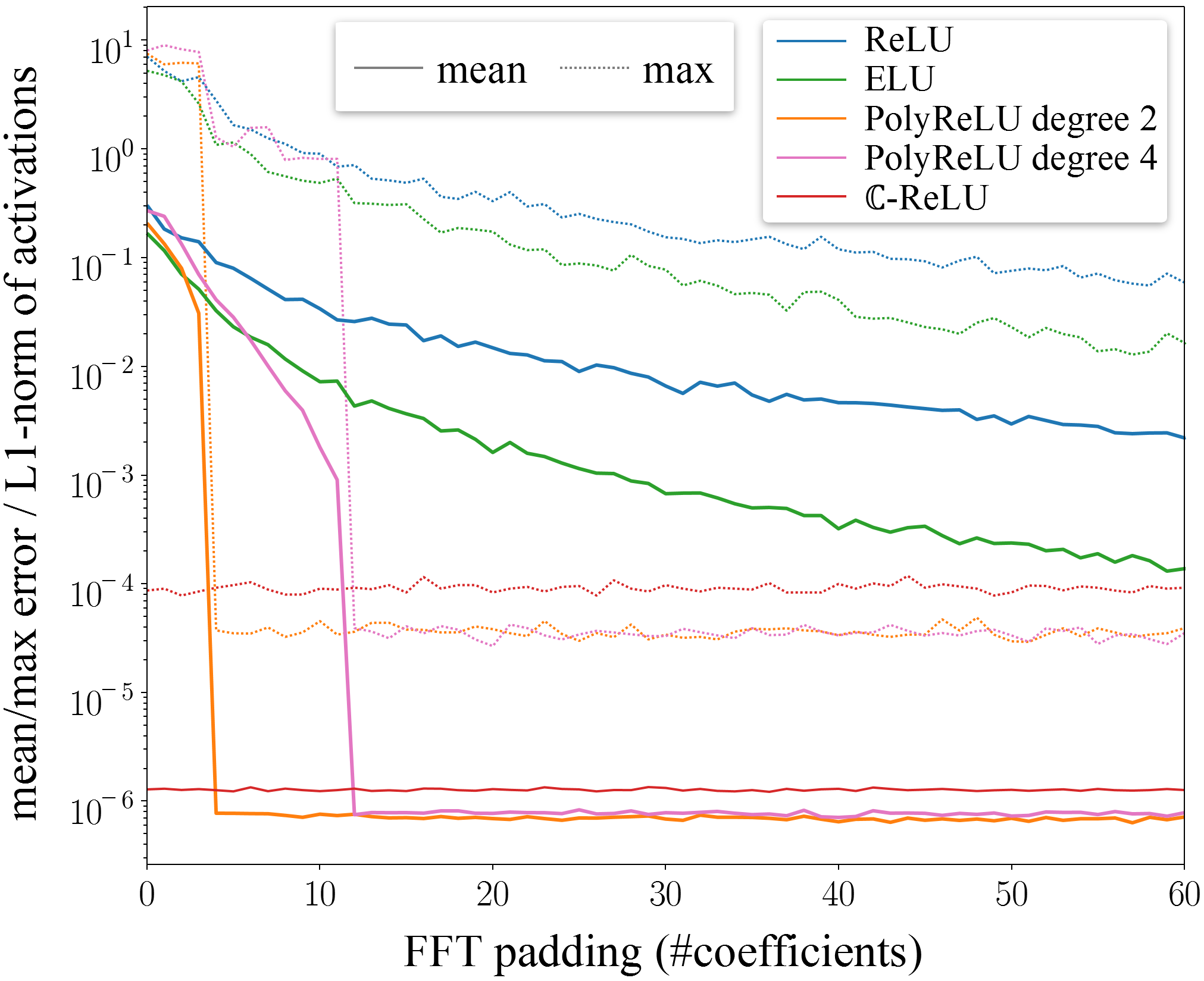} 
			\caption{Same errors as in Fig.~\ref{fig:errorReLU} after the fifth (penultimate) equivariant layer for various nonlinearities. Polynomials show the expected sharp decline at with increasing FFT padding. $\doubleC$-$\ReLU$ included as reference, (no FFT used). Note the different scale on the axes.}
			\label{fig:errorVarious}
	\end{minipage}}
\end{figure}

We have implemented the $SO(2)$-equivariant layers and the corresponding $SE(2)$- and $SE(3)$-equivariant point cloud / surfel networks in PyTorch using PyKeOps~\cite{PyKeOps} for computing the sparse matrix-vector multiplications of the general point clouds efficiently. The source code is provided as supplementary material.

\begin{table}
  \caption{Results on the {\mnistrot} dataset}
  \label{tab:mnist}
  \centering
	\scriptsize
  \begin{tabular}{ccccccccccc}
    \toprule
          &                & repre-    & num.   & FFT & activation         & invariant        & model     & sec / & \multicolumn{2}{c}{test error (\%)} \\
    model & group          & sentation & coeff. & pad & function           & map              & param.    & epoch & mean  & std   \\
		\midrule
    E2CNN~\cite{Weiler-generalE2-NeurIPS19} & $C_{16}$         & regular   & 16 & - & $\ELU$ & \emph{maxpool} & 2,692,690 & 38 & 0.716 & 0.028 \\
    E2CNN~\cite{Weiler-generalE2-NeurIPS19} & $C_{16}$         & quotient  & 16 & - & $\ELU$ & \emph{maxpool} & 2,749,686 & 49 & 0.705 & 0.045 \\
    E2CNN~\cite{Weiler-generalE2-NeurIPS19} & $D_{16|5}C_{16}$ & regular   & 16 & - & $\ELU$ & \emph{maxpool} & 3,069,353 & 76 & 0.682 & 0.022 \\
		\midrule
		Ours  & $SO(2)$        & Fourier   &  9     &   - & $\doubleC$-$\ReLU$ & \emph{norm}      & 1,396,138 & 30   & 0.980 & 0.031 \\
		Ours  & $SO(2)$        & Fourier   &  9     & 127 & $\ELU$             & \emph{norm}      & 1,394,986 & 36   & 0.729 & 0.029 \\
		Ours  & $SO(2)$        & Fourier   &  9     & 127 & $\tanh$            & \emph{norm}      & 1,394,986 & 36   & 0.768 & 0.024 \\
		Ours  & $SO(2)$        & Fourier   &  9     & 127 & $\ReLU$            & \emph{norm}      & 1,394,986 & 36   & 0.685 & 0.026 \\
		Ours  & $SO(2)$        & Fourier   &  9     &   7 & $\ReLU$            & \emph{norm}      & 1,394,986 & 30   & 0.689 & 0.019 \\
		Ours  & $SO(2)$        & Fourier   & 17     & 127 & $\ReLU$            & \emph{norm}      & 2,729,098 & 64   & 0.699 & 0.033 \\
		Ours  & $SO(2)$        & Fourier   &  9     & 127 & $\ReLU$            & \emph{conv2triv} &   891,178 & 36   & 0.719 & 0.018 \\
		\midrule
		Ours  & $SO(2)$        & Fourier   &  9     &   8 & Poly(2)            & \emph{norm}      & 1,394,986 & 32   & 0.690 & 0.015 \\
		Ours  & $SO(2)$        & Fourier   &  9     &  24 & Poly(4)            & \emph{norm}      & 1,394,986 & 36   & 0.690 & 0.024 \\
    \bottomrule
  \end{tabular}
\end{table}

\begin{table}
  \caption{Results on the {\modelnetforty} dataset}
  \label{tab:modelnet40}
  \centering
	\scriptsize
  \begin{tabular}{llccc}
    \toprule
                                        &                        & \multicolumn{3}{c}{test accuracy (\%)}  \\ \cmidrule(r){3-5}
    method                              & training time          & N/$SO(3)$ & $z$/$SO(3)$  & $SO(3)$/$SO(3)$ \\
		\midrule
		Spherical CNNs~\cite{SphericalCNNs} & -                      & -         & -            & 81.3 \\
		SPHnet~\cite{SPHnet}                & 1.5 hrs (RTX 2080 Ti)  & 86.6      & -            & 87.6 \\
		CSGNN~\cite{CSGNN}                  & 6.8 hrs (Tesla P100)   & -         & 86.2         & 88.9 \\
		MA-KPConv~\cite{MAKPConv}           & 7.5 hrs (Tesla  V100)  & 89.1      & 89.1         & 89.1 \\
		GCANet~\cite{GCANet}                & -                      & -         & 89.1         & 89.2 \\
		\midrule
		Ours (Poly(2), FFT pad 8)           & 55 min (RTX 2080 Ti) (w/o preprocessing)  & 88.0      & 88.0         & 88.0 \\
		Ours (Poly(4), FFT pad 24)          & 60 min (RTX 2080 Ti) (w/o preprocessing)  & 88.7      & 88.7         & 88.7 \\
		Ours ($\ReLU$, FFT pad 127)         & 60 min (RTX 2080 Ti) (w/o preprocessing)  & 89.1      & 89.1         & 89.1 \\
    \bottomrule
  \end{tabular}
\end{table}

\textbf{Data sets:} We test our implementation on image and 3D data. In the image case, we replicate the architecture used by \citet{Weiler-generalE2-NeurIPS19} on the {\mnistrotlink} dataset from their recent survey of $E(2)$-equivariant image processing networks. For simplicity, we convert all image data into point clouds. This comes with some preprocessing overhead, but the absolute training times are still comparable to the pixel-based implementation used in \cite{Weiler-generalE2-NeurIPS19}, see Table~\ref{tab:mnist}. The point-based representation can be rotated exactly, facilitating the measurement of accuracy in terms of equivariance (images require supersampling to the remove aliasing of the pixel grid after rotations).

For the 3D surfel case, we use {\modelnetfortylink} as benchmark. We rescale all models to a unit bounding cube and convert the polygonal data into point clouds by z-Buffer rasterization from 50 random view points. Normals are estimated from a PCA-fit to 20 nearest neighbors at a sample spacing of $0.005$, and oriented to point away from the center of mass. The final input is obtained by a reduction by Poisson-disc sampling with a sample spacing of \ok{$0.05$}.

\textbf{Accuracy of Equivariance}
Figure~\ref{fig:errorReLU} shows the error for the FFT-sampled $\ReLU$ for various amount of oversampling (padding applied to the Fourier basis) for all layers of the {\mnistrot}-network. The different equivariant layers show a similar relative error, with the error on the final invariant (logits) layer being lower. All experiments use 32-bit floating-point GPU computations.

We compare the error approximations of different nonlinearities in Figure~\ref{fig:errorVarious}. The error for the polynomial nonlinearities drops sharply when a specific amount of oversampling is applied, which is in accordance with our theoretical considerations from Section~\ref{sec:polyfft}. From this point on, further oversampling does not improve equivariance, which suggests that the remaining small fluctuations are due to the numerical limitations of the 32-bit floating point representation. This is supported by the observation that $\doubleC$-$\ReLU$, which should be perfectly equivariant as it only operates on the norm of the Fourier coefficients, produces a similar level of fluctuations.

The error for the approximations of ReLU and ELU drop continuously with increasing oversampling, with ELU dropping significantly faster that ReLU. The convergence behavior is good enough to reach reasonable accuracies (mean output errors of $10^{-5}$) with feasible oversampling (up to 512 coefficients). For high accuracy requirements, the polynomial approach appears to be favorable (with an order of magnitude higher accuracy, at the limits of 32-bit floating point).
%However, both do not reach the same level of equivariance as the polynomial functions even when applying a large oversampling to the FFT.

\textbf{Prediction Accuracy:}
Table~\ref{tab:mnist} lists our results on the {\mnistrot} dataset, together with those obtained by \citet{Weiler-generalE2-NeurIPS19}, which we managed to reproduce using their published code. Our models using $\ReLU$ or its polynomial approximations reach comparable performance to the best rotation-only ($C_{16}$) equivariant models evaluated this paper, while having a lower overall parameter count. Interestingly, the amount of oversampling (if we compare the $\ReLU$-models with padding 7 and 127) does not seem to have a large impact on the accuracy score, while models using the $\ELU$ or $\tanh$ nonlinearities produce slightly worse results.

The results on {\modelnetforty} are compared to the literature in Table~\ref{tab:modelnet40}. As common in the literature, we refer to a model trained on the original (non-augmented) dataset and evaluated with random rotations as N/$SO(3)$, while we denote rotational augmentation during training and testing as $SO(3)$/$SO(3)$. We also include a benchmarks for z/$SO(3)$, denoting random rotations around the $z$-axis during training and full rotations during testing, which is used in some papers. We performed the evaluation for all 3 regimes to allow a fair comparison. The resulting accuracy of our surfel based model compares favorable with various other architecture. When using the $\ReLU$ nonlinearity, results are on par with other state-of-the-art architecture on this task, while polynominal activation functions perform slightly worse.

\textbf{Computational Cost}
In Table~\ref{tab:mnist}, we list training time in seconds per epoch obtained on a single \emph{Nvidia RTX 2080Ti} graphics card. We can estimate the compuational cost of the FFT by comparing the runtimes with those of the $\doubleC$-$\ReLU$, which can be quickly computed and does not require performing an FFT. This suggests the the overhead of computing the FFT is low (smaller than 20\%, depending on the amount of oversampling) compared to the other operations performed. Note that the times per epoch given in Table~\ref{tab:mnist} also reflect total training time, as all models (E2CNN~\cite{Weiler-generalE2-NeurIPS19} and Ours) are trained for 40 epochs.

\section{Conclusions \& Future Work}

In this paper, we have presented an analysis of the effect of nonlinear activation functions on the Fourier representations used by $SO(2)$-equivariant networks. The main insight is that the nonlinearity creates high frequency harmonics; thus applying the nonlinearity to an oversampled angular domain representation can maintain equivariance. For polynomial nonlinearities, this construction is provably exact. This theoretical prediction is also observed in real numerical implementation. In the general case, we empirically observe rapid convergence, which is also plausible from an analytical perspective. As a sanity check, we have applied our method to shape and image classification, and reach performance on par with state-of-the-art equivariant architectures while providing full, continuous equivariance.

Our main result, the oversampling algorithm for applying nonlinearities closes a small, but important gap in the literature on equivariant networks for iamge and geometric data processing: It removes most architectural restictions, making the design of such networks significantly easier. The algorithm employed is easy to understand and implement; the only (but, as our experiments show, crucial) departure from base-line angular-domain evaluation is a resolution increase.

The method provides continuous equivariance up to numerical precision. In crictical applications, such as physical simulations, polynomial non-linearities can optionally provide an a priori guarantee; however, with empirical calibration of oversampling factor $D$, general non-linearities can be used. In both cases, when integrated into a point cloud network, the performance penalty is very minor in relation to the overall costs.

The biggest conceptual limitation of this approach is probably the restriction to $SO(2)$. While similar construction could probably be applied to more complex compact Lie groups like $SO(3)$ (without using auxiliary orientation information), performing harmonic distortion analysis in non-commutative symmetry groups becomes significantly more involved (requiring for example a matrix-valued convolution theorem).

\section*{Acknowledgements}
This work was supported by the Collaborative Research Center TRR 146 of the Deutsche Forschungsgemeinschaft (DFG).

%\section*{References}

{
\small
\bibliography{nonlinearities-s2021}

\begin{thebibliography}{37}
\providecommand{\natexlab}[1]{#1}
\providecommand{\url}[1]{\texttt{#1}}
\expandafter\ifx\csname urlstyle\endcsname\relax
  \providecommand{\doi}[1]{doi: #1}\else
  \providecommand{\doi}{doi: \begingroup \urlstyle{rm}\Url}\fi

\bibitem[{Ali Mehmeti-Göpel} et~al.(2021){Ali Mehmeti-Göpel}, Hartmann, and
  Wand]{RingingReLUs-AliMehmetiGoepel-2021}
C.~H. {Ali Mehmeti-Göpel}, D.~Hartmann, and M.~Wand.
\newblock Ringing {ReLUs}: Harmonic distortion analysis of nonlinear
  feedforward networks.
\newblock In \emph{International Conference on Learning Representations
  (ICLR).}, 2021.
\newblock URL \url{https://openreview.net/forum?id=TaYhv-q1Xit}.

\bibitem[Atzmon et~al.(2018)Atzmon, Maron, and Lipman]{PCNN}
M.~Atzmon, H.~Maron, and Y.~Lipman.
\newblock Point convolutional neural networks by extension operators.
\newblock \emph{ACM Transactions on Graphics (TOG)}, 37:\penalty0 1 -- 12,
  2018.

\bibitem[Charlier et~al.(2021)Charlier, Feydy, Glaunès, Collin, and
  Durif]{PyKeOps}
B.~Charlier, J.~Feydy, J.~A. Glaunès, F.-D. Collin, and G.~Durif.
\newblock Kernel operations on the gpu, with autodiff, without memory
  overflows.
\newblock \emph{Journal of Machine Learning Research}, 22\penalty0
  (74):\penalty0 1--6, 2021.
\newblock URL \url{http://jmlr.org/papers/v22/20-275.html}.

\bibitem[Clevert et~al.(2016)Clevert, Unterthiner, and
  Hochreiter]{clevert2016fast}
D.-A. Clevert, T.~Unterthiner, and S.~Hochreiter.
\newblock Fast and accurate deep network learning by exponential linear units
  (elus), 2016.

\bibitem[Cohen and Welling(2016)]{Cohen-GCNN-ICML-2016}
T.~S. Cohen and M.~Welling.
\newblock Group equivariant convolutional networks.
\newblock In \emph{Proceedings of the International Conference on Machine
  Learning (ICML)}, 2016.

\bibitem[Cohen and Welling(2017)]{Cohen-Steerable-2017}
T.~S. Cohen and M.~Welling.
\newblock Steerable cnns.
\newblock In \emph{International Conference on Learning Representations
  (ICLR)}, 2017.

\bibitem[Dieleman et~al.(2016)Dieleman, Fauw, and
  Kavukcuoglu]{dieleman2016exploiting}
S.~Dieleman, J.~D. Fauw, and K.~Kavukcuoglu.
\newblock Exploiting cyclic symmetry in convolutional neural networks, 2016.

\bibitem[Esteves et~al.(2017)Esteves, Allen-Blanchette, Makadia, and
  Daniilidis]{SphericalCNNs}
C.~Esteves, C.~Allen-Blanchette, A.~Makadia, and K.~Daniilidis.
\newblock 3d object classification and retrieval with spherical cnns.
\newblock \emph{ArXiv}, abs/1711.06721, 2017.

\bibitem[Feynman et~al.(1965)Feynman, Leighton, and
  Sands]{feynman-lectures-1965}
R.~P. Feynman, R.~B. Leighton, and M.~Sands.
\newblock The feynman lectures on physics; vol. i.
\newblock \emph{American Journal of Physics}, 50\penalty0 (8), 1965.

\bibitem[Fox et~al.(2021)Fox, Zhao, Rajamanickam, Ramprasad, and Song]{CSGNN}
J.~Fox, B.~Zhao, S.~Rajamanickam, R.~Ramprasad, and L.~Song.
\newblock Concentric spherical {GNN} for 3d representation learning.
\newblock \emph{CoRR}, abs/2103.10484, 2021.
\newblock URL \url{https://arxiv.org/abs/2103.10484}.

\bibitem[Glassner(1995)]{Glassner-ImageSynthesis-1995}
A.~S. Glassner.
\newblock \emph{Principles of Digital Image Synthesis}.
\newblock Morgan Kaufmann Publishers, 1995.

\bibitem[Gottemukkula(2020)]{Gottemukkula2020PolynomialActications}
V.~Gottemukkula.
\newblock Polynomial activation functions.
\newblock In \emph{International Conference on Learning Representations
  (retracted paper)}, 2020.
\newblock URL \url{https://openreview.net/forum?id=rkxsgkHKvH}.

\bibitem[Hanocka et~al.(2019)Hanocka, Hertz, Fish, Giryes, Fleishman, and
  Cohen-Or]{MeshCNN}
R.~Hanocka, A.~Hertz, N.~Fish, R.~Giryes, S.~Fleishman, and D.~Cohen-Or.
\newblock Meshcnn: a network with an edge.
\newblock \emph{ACM Transactions on Graphics (TOG)}, 38:\penalty0 1 -- 12,
  2019.

\bibitem[He et~al.(2015)He, Zhang, Ren, and Sun]{HeInit}
K.~He, X.~Zhang, S.~Ren, and J.~Sun.
\newblock Delving deep into rectifiers: Surpassing human-level performance on
  imagenet classification.
\newblock In \emph{Proceedings of the 2015 IEEE International Conference on
  Computer Vision (ICCV)}, ICCV '15, page 1026–1034, USA, 2015. IEEE Computer
  Society.

\bibitem[Hoogeboom et~al.(2018)Hoogeboom, Peters, Cohen, and
  Welling]{hoogeboom2018hexaconv}
E.~Hoogeboom, J.~W.~T. Peters, T.~S. Cohen, and M.~Welling.
\newblock Hexaconv, 2018.

\bibitem[Ioffe and Szegedy(2015)]{Batchnorm-Ioffe-ICML-2015}
S.~Ioffe and C.~Szegedy.
\newblock Batch normalization: Accelerating deep network training by reducing
  internal covariate shift.
\newblock In F.~R. Bach and D.~M. Blei, editors, \emph{Proceedings of the 32nd
  International Conference on Machine Learning, {ICML} 2015, Lille, France,
  6-11 July 2015}, volume~37 of \emph{{JMLR} Workshop and Conference
  Proceedings}, pages 448--456. JMLR.org, 2015.

\bibitem[Kingma and Ba(2015)]{Kingma2015AdamAM}
D.~P. Kingma and J.~Ba.
\newblock Adam: A method for stochastic optimization.
\newblock \emph{CoRR}, abs/1412.6980, 2015.

\bibitem[Kondor et~al.(2018)Kondor, Hy, Pan, Anderson, and
  Trivedi]{Kondor-covariantgraph-ICLR-2018}
R.~Kondor, T.~S. Hy, H.~Pan, B.~M. Anderson, and S.~Trivedi.
\newblock Covariant compositional networks for learning graphs.
\newblock In \emph{ICLR 2018 Workshop Track}, 2018.
\newblock URL \url{https://openreview.net/forum?id=S1TgE7WR-}.

\bibitem[Laptev et~al.(2016)Laptev, Savinov, Buhmann, and
  Pollefeys]{laptev2016tipooling}
D.~Laptev, N.~Savinov, J.~M. Buhmann, and M.~Pollefeys.
\newblock Ti-pooling: transformation-invariant pooling for feature learning in
  convolutional neural networks, 2016.

\bibitem[Li et~al.(2018)Li, Bu, Sun, Wu, Di, and Chen]{PointCNN}
Y.~Li, R.~Bu, M.~Sun, W.~Wu, X.~Di, and B.~Chen.
\newblock Pointcnn: Convolution on {X}-transformed points.
\newblock \emph{arXiv: Computer Vision and Pattern Recognition}, 2018.

\bibitem[Pfister et~al.(2000)Pfister, Zwicker, van Baar, and
  Gross]{Pfister-Surfels-2000}
H.~Pfister, M.~Zwicker, J.~van Baar, and M.~Gross.
\newblock Surfels: Surface elements as rendering primitives.
\newblock In \emph{ACM SIGGRAPH 2000 Conference Proceedings, Annual Conference
  Series}, 2000.

\bibitem[Poulenard et~al.(2019)Poulenard, Rakotosaona, Ponty, and
  Ovsjanikov]{SPHnet}
A.~Poulenard, M.-J. Rakotosaona, Y.~Ponty, and M.~Ovsjanikov.
\newblock Effective rotation-invariant point cnn with spherical harmonics
  kernels.
\newblock \emph{2019 International Conference on 3D Vision (3DV)}, pages
  47--56, 2019.

\bibitem[Qi et~al.(2017{\natexlab{a}})Qi, Su, Mo, and Guibas]{PointNet}
C.~Qi, H.~Su, K.~Mo, and L.~Guibas.
\newblock Pointnet: Deep learning on point sets for 3d classification and
  segmentation.
\newblock \emph{2017 IEEE Conference on Computer Vision and Pattern Recognition
  (CVPR)}, pages 77--85, 2017{\natexlab{a}}.

\bibitem[Qi et~al.(2017{\natexlab{b}})Qi, Yi, Su, and Guibas]{PointNetPP}
C.~Qi, L.~Yi, H.~Su, and L.~Guibas.
\newblock Pointnet++: Deep hierarchical feature learning on point sets in a
  metric space.
\newblock In \emph{NIPS}, 2017{\natexlab{b}}.

\bibitem[Sch{\"u}tt et~al.(2017)Sch{\"u}tt, Kindermans, Felix, Chmiela,
  Tkatchenko, and M{\"u}ller]{SchNet}
K.~Sch{\"u}tt, P.-J. Kindermans, H.~E.~S. Felix, S.~Chmiela, A.~Tkatchenko, and
  K.~M{\"u}ller.
\newblock Schnet: A continuous-filter convolutional neural network for modeling
  quantum interactions.
\newblock In \emph{NIPS}, 2017.

\bibitem[Srivastava et~al.(2014)Srivastava, Hinton, Krizhevsky, Sutskever, and
  Salakhutdinov]{Srivastava2014DropoutAS}
N.~Srivastava, G.~E. Hinton, A.~Krizhevsky, I.~Sutskever, and R.~Salakhutdinov.
\newblock Dropout: a simple way to prevent neural networks from overfitting.
\newblock \emph{J. Mach. Learn. Res.}, 15:\penalty0 1929--1958, 2014.

\bibitem[Su et~al.(2015)Su, Maji, Kalogerakis, and
  Learned-Miller]{Su2015MultiviewCN}
H.~Su, S.~Maji, E.~Kalogerakis, and E.~Learned-Miller.
\newblock Multi-view convolutional neural networks for 3d shape recognition.
\newblock \emph{2015 IEEE International Conference on Computer Vision (ICCV)},
  pages 945--953, 2015.

\bibitem[Thomas(2020)]{MAKPConv}
H.~Thomas.
\newblock Rotation-invariant point convolution with multiple equivariant
  alignments.
\newblock \emph{2020 International Conference on 3D Vision (3DV)}, pages
  504--513, 2020.

\bibitem[Thomas et~al.(2019)Thomas, Qi, Deschaud, Marcotegui, Goulette, and
  Guibas]{KPConv}
H.~Thomas, C.~Qi, J.-E. Deschaud, B.~Marcotegui, F.~Goulette, and L.~Guibas.
\newblock Kpconv: Flexible and deformable convolution for point clouds.
\newblock \emph{2019 IEEE/CVF International Conference on Computer Vision
  (ICCV)}, pages 6410--6419, 2019.

\bibitem[Thomas et~al.(2018)Thomas, Smidt, Kearnes, Yang, Li, Kohlhoff, and
  Riley]{TensorFieldNetworks}
N.~Thomas, T.~Smidt, S.~M. Kearnes, L.~Yang, L.~Li, K.~Kohlhoff, and P.~Riley.
\newblock Tensor field networks: Rotation- and translation-equivariant neural
  networks for 3d point clouds.
\newblock \emph{CoRR}, abs/1802.08219, 2018.
\newblock URL \url{http://arxiv.org/abs/1802.08219}.

\bibitem[Weiler and Cesa(2019)]{Weiler-generalE2-NeurIPS19}
M.~Weiler and G.~Cesa.
\newblock General {E(2)}-equivariant steerable cnns.
\newblock In H.~Wallach, H.~Larochelle, A.~Beygelzimer, F.~d\textquotesingle
  Alch\'{e}-Buc, E.~Fox, and R.~Garnett, editors, \emph{Advances in Neural
  Information Processing Systems}, volume~32, pages 14334--14345. Curran
  Associates, Inc., 2019.
\newblock URL
  \url{https://proceedings.neurips.cc/paper/2019/file/45d6637b718d0f24a237069fe41b0db4-Paper.pdf}.

\bibitem[Weiler et~al.(2018{\natexlab{a}})Weiler, Boomsma, Geiger, Welling, and
  Cohen]{Cohen-3DSteerable-2018}
M.~Weiler, W.~Boomsma, M.~Geiger, M.~Welling, and T.~S. Cohen.
\newblock 3d steerable cnns, learning rotationally equivariant features in
  volumetric data.
\newblock In \emph{Advances in Neural Information Processing Systems
  (NeurIPS)}, 2018{\natexlab{a}}.

\bibitem[Weiler et~al.(2018{\natexlab{b}})Weiler, Hamprecht, and
  Storath]{weiler2018learning}
M.~Weiler, F.~A. Hamprecht, and M.~Storath.
\newblock Learning steerable filters for rotation equivariant cnns,
  2018{\natexlab{b}}.

\bibitem[Wiersma et~al.(2020)Wiersma, Eisemann, and
  Hildebrandt]{Wiersma-SurfaceCNNs-Siggraph-2020}
R.~Wiersma, E.~Eisemann, and K.~Hildebrandt.
\newblock {CNNs} on surfaces using rotation-equivariant features.
\newblock \emph{Transactions on Graphics}, 39\penalty0 (4), July 2020.
\newblock \doi{10.1145/3386569.3392437}.

\bibitem[Worrall et~al.(2017)Worrall, Garbin, Turmukhambetov, and
  Brostow]{worrall2017harmonic}
D.~E. Worrall, S.~J. Garbin, D.~Turmukhambetov, and G.~J. Brostow.
\newblock Harmonic networks: Deep translation and rotation equivariance, 2017.

\bibitem[Wu et~al.(2015)Wu, Song, Khosla, Yu, Zhang, Tang, and Xiao]{ModelNet}
Z.~Wu, S.~Song, A.~Khosla, F.~Yu, L.~Zhang, X.~Tang, and J.~Xiao.
\newblock 3d shapenets: A deep representation for volumetric shapes.
\newblock In \emph{2015 IEEE Conference on Computer Vision and Pattern
  Recognition (CVPR)}, pages 1912--1920, 2015.
\newblock \doi{10.1109/CVPR.2015.7298801}.

\bibitem[Zhang et~al.(2020)Zhang, Hua, Chen, Tian, and Yeung]{GCANet}
Z.~Zhang, B.-S. Hua, W.~Chen, Y.~Tian, and S.~Yeung.
\newblock Global context aware convolutions for 3d point cloud understanding.
\newblock \emph{2020 International Conference on 3D Vision (3DV)}, pages
  210--219, 2020.

\end{thebibliography}


\begin{thebibliography}{9}
\providecommand{\natexlab}[1]{#1}
\providecommand{\url}[1]{\texttt{#1}}
\expandafter\ifx\csname urlstyle\endcsname\relax
  \providecommand{\doi}[1]{doi: #1}\else
  \providecommand{\doi}{doi: \begingroup \urlstyle{rm}\Url}\fi

\bibitem[Gottemukkula(2020)]{Gottemukkula2020PolynomialActications}
V.~Gottemukkula.
\newblock Polynomial activation functions.
\newblock In \emph{International Conference on Learning Representations
  (retracted paper)}, 2020.
\newblock URL \url{https://openreview.net/forum?id=rkxsgkHKvH}.

\bibitem[He et~al.(2015)He, Zhang, Ren, and Sun]{HeInit}
K.~He, X.~Zhang, S.~Ren, and J.~Sun.
\newblock Delving deep into rectifiers: Surpassing human-level performance on
  imagenet classification.
\newblock In \emph{Proceedings of the 2015 IEEE International Conference on
  Computer Vision (ICCV)}, ICCV '15, page 1026–1034, USA, 2015. IEEE Computer
  Society.

\bibitem[Ioffe and Szegedy(2015)]{Batchnorm-Ioffe-ICML-2015}
S.~Ioffe and C.~Szegedy.
\newblock Batch normalization: Accelerating deep network training by reducing
  internal covariate shift.
\newblock In F.~R. Bach and D.~M. Blei, editors, \emph{Proceedings of the 32nd
  International Conference on Machine Learning, {ICML} 2015, Lille, France,
  6-11 July 2015}, volume~37 of \emph{{JMLR} Workshop and Conference
  Proceedings}, pages 448--456. JMLR.org, 2015.

\bibitem[Srivastava et~al.(2014)Srivastava, Hinton, Krizhevsky, Sutskever, and
  Salakhutdinov]{Srivastava2014DropoutAS}
N.~Srivastava, G.~E. Hinton, A.~Krizhevsky, I.~Sutskever, and R.~Salakhutdinov.
\newblock Dropout: a simple way to prevent neural networks from overfitting.
\newblock \emph{J. Mach. Learn. Res.}, 15:\penalty0 1929--1958, 2014.

\bibitem[Weiler and Cesa(2019)]{Weiler-generalE2-NeurIPS19}
M.~Weiler and G.~Cesa.
\newblock General {E(2)}-equivariant steerable cnns.
\newblock In H.~Wallach, H.~Larochelle, A.~Beygelzimer, F.~d\textquotesingle
  Alch\'{e}-Buc, E.~Fox, and R.~Garnett, editors, \emph{Advances in Neural
  Information Processing Systems}, volume~32, pages 14334--14345. Curran
  Associates, Inc., 2019.
\newblock URL
  \url{https://proceedings.neurips.cc/paper/2019/file/45d6637b718d0f24a237069fe41b0db4-Paper.pdf}.

\bibitem[Wiersma et~al.(2020)Wiersma, Eisemann, and
  Hildebrandt]{Wiersma-SurfaceCNNs-Siggraph-2020}
R.~Wiersma, E.~Eisemann, and K.~Hildebrandt.
\newblock {CNNs} on surfaces using rotation-equivariant features.
\newblock \emph{Transactions on Graphics}, 39\penalty0 (4), July 2020.
\newblock \doi{10.1145/3386569.3392437}.

\bibitem[Worrall et~al.(2017)Worrall, Garbin, Turmukhambetov, and
  Brostow]{worrall2017harmonic}
D.~E. Worrall, S.~J. Garbin, D.~Turmukhambetov, and G.~J. Brostow.
\newblock Harmonic networks: Deep translation and rotation equivariance, 2017.

\bibitem[Wu et~al.(2015)Wu, Song, Khosla, Yu, Zhang, Tang, and Xiao]{ModelNet}
Z.~Wu, S.~Song, A.~Khosla, F.~Yu, L.~Zhang, X.~Tang, and J.~Xiao.
\newblock 3d shapenets: A deep representation for volumetric shapes.
\newblock In \emph{2015 IEEE Conference on Computer Vision and Pattern
  Recognition (CVPR)}, pages 1912--1920, 2015.
\newblock \doi{10.1109/CVPR.2015.7298801}.

\bibitem[Yuksel(2015)]{Yuksel2015}
C.~Yuksel.
\newblock Sample elimination for generating poisson disk sample sets.
\newblock \emph{Computer Graphics Forum (Proceedings of EUROGRAPHICS 2015)},
  34\penalty0 (2):\penalty0 25--32, 2015.
\newblock ISSN 0167-7055.
\newblock \doi{10.1111/cgf.12538}.
\newblock URL \url{http://dx.doi.org/10.1111/cgf.12538}.

\end{thebibliography}
}

\end{document}

\maketitle

\appendix

\section{Supplementary Material}

In this appendix, we will give detailed information about our architectures and training procedures and show additional results that we could not fit in the main paper.

\subsection{$SE(2)$-Equivariant Networks for 2D Point Clouds}
\label{sec:appendix_mnist}

First, we would like to explain more in detail how we performed the experiments comparing the 2D networks to the benchmark reported by~\citet{Weiler-generalE2-NeurIPS19}.

For simplicity, our 2D implementation operates directly on point clouds (this is not a fundamental limitation; as the point cloud code is more general, we can use it to handle pixel grids, too). As the {\mnistrotlink} dataset consists of images, we first convert them to point clouds by placing the values on a regular grid with a spacing of $1$, i.e. we generate a \emph{coordinate} tensor containing the point coordinates and a \emph{data} tensor containing the pixel values for each pixel in the image. The \emph{data} tensor of the input is real-valued with rotation order~$0$, but in general, its values in the equivariant part of the network are complex Fourier coefficients of various rotation order, representing the network activations as band-limited angular functions.

\textbf{Architecture:}
Architecture-wise, we stay as close as possible to the best performing rotation equivariant {\mnistrot} architecture by~\citet{Weiler-generalE2-NeurIPS19}, i.e. we use the same number, type and width of layers, as well as identical filter parameters for the equivariant convolutional layers. A detailed view of our architecture is shown in table~\ref{tab:detailed_mnist_arch}.

Our network consists of 6 rotation-equivariant convolutional layers, after which an \emph{invariance operation} is applied, followed by 3 rotation-invariant linear (fully-connected) layers. After each layer, we perform Batch Normalization~\cite{Batchnorm-Ioffe-ICML-2015}, followed by a nonlinearity, with exception of the final linear layer. When working on complex Fourier coefficients, we apply a custom Batch Normalization layer which operates directly on the coefficients (as discussed in the main paper), otherwise, we use the standard implementation from \emph{PyTorch}. Futhermore, we also apply Dropout~\cite{Srivastava2014DropoutAS} with $p=0.3$ to the inputs of each linear layer.

Contrary to \citet{Weiler-generalE2-NeurIPS19}, who use spatial max-pooling in their best-performing architectures, we opt for $2*2$ spatial average pooling instead, which has the advantage that it can be applied directly on the Fourier coefficients without requiring any transformation. After each pooling step, we divide the \emph{coordinate} tensor by $2$, so the spacing between grid points is kept at $1$ at all times.

\textbf{Training and testing:}
We train out networks on the training set of {\mnistrotlink}, containing 12,000 rotated images from the original MNIST dataset, and evaluate them on the corresponding test set, containing 50,000 images, using the same data loading pipeline and hyperparameters as~\citet{Weiler-generalE2-NeurIPS19}. Network weights are initialized randomly using He initialization~\cite{HeInit}, and then trained using the Adam optimizer with \emph{PyTorch} default parameters ($\textrm{betas}=(0.9, 0.999)$, $\textrm{eps}=10^{-8}$) and a mini-batch size of 64 to optimize the log-softmax cross entropy of the network output. Training is done for a total of 40 epochs, using a fixed learning rate of 0.015 in the first 16 epochs, an then decayed exponentially by applying a factor of 0.8 before each following epoch. To get meaningful results, we calculate the mean test error and its standard deviation from 10 independent training runs. 

Between training and testing, we perform an additional pass on the training set without changing trainable weights to calculate the exact statistics on the full training set in the Batch Normalization layers. This gives more representative values than using an expontial average for determining the mean and variance values during the optimization process.

\begin{table}
  \caption{Detailed architecture for experiments on the {\mnistrot} dataset}
  \centering
	\scriptsize
  \begin{tabular}{cccccc}
\toprule
\multirow{2}{*}{no.} & layer                 & spatial                  & output size                                  & filters                  & \multirow{2}{*}{followed by} \\
										 & type                  & reduction                & $x*y*\textrm{channels}*\textrm{coeffs}$      & ($r$, $\sigma$, frequencies) & \\
\midrule
								0  &                 Input   &                 -        &                  $28*28*1*1               $  & -                        & -- \\
\midrule
\multirow{5}{*}{1} & \multirow{5}{*}{Conv2D} & \multirow{5}{*}{crop(4)} & \multirow{5}{*}{$20*20*24*c_\textrm{max}  $} & ($0$, $0.005$, $0     $) & \multirow{5}{*}{BN + nonlinearity} \\
                   &                         &                          &                                              & ($1$, $0.600$, $-2...2$) \\
                   &                         &                          &                                              & ($2$, $0.600$, $-3...3$) \\
                   &                         &                          &                                              & ($3$, $0.600$, $-6...6$) \\
                   &                         &                          &                                              & ($4$, $0.400$, $-2...2$) \\ 
\addlinespace[0.5em]
\multirow{4}{*}{2} & \multirow{4}{*}{Conv2D} & \multirow{4}{*}{pool(2)} & \multirow{4}{*}{$10*10*32*c_\textrm{max}  $} & ($0$, $0.005$, $0     $) & \multirow{4}{*}{BN + nonlinearity} \\
                   &                         &                          &                                              & ($1$, $0.600$, $-2...2$) \\
                   &                         &                          &                                              & ($2$, $0.600$, $-3...3$) \\
                   &                         &                          &                                              & ($3$, $0.400$, $-2...2$) \\
\addlinespace[0.5em]
\multirow{4}{*}{3} & \multirow{4}{*}{Conv2D} & \multirow{4}{*}{--     } & \multirow{4}{*}{$10*10*36*c_\textrm{max}  $} & ($0$, $0.005$, $0     $) & \multirow{4}{*}{BN + nonlinearity} \\
                   &                         &                          &                                              & ($1$, $0.600$, $-2...2$) \\
                   &                         &                          &                                              & ($2$, $0.600$, $-3...3$) \\
                   &                         &                          &                                              & ($3$, $0.400$, $-2...2$) \\
\addlinespace[0.5em]
\multirow{4}{*}{4} & \multirow{4}{*}{Conv2D} & \multirow{4}{*}{pool(2)} & \multirow{4}{*}{$ 5* 5*36*c_\textrm{max}  $} & ($0$, $0.005$, $0     $) & \multirow{4}{*}{BN + nonlinearity} \\
                   &                         &                          &                                              & ($1$, $0.600$, $-2...2$) \\
                   &                         &                          &                                              & ($2$, $0.600$, $-3...3$) \\
                   &                         &                          &                                              & ($3$, $0.400$, $-2...2$) \\
\addlinespace[0.5em]
\multirow{4}{*}{5} & \multirow{4}{*}{Conv2D} & \multirow{4}{*}{--     } & \multirow{4}{*}{$ 5* 5*64*c_\textrm{max}  $} & ($0$, $0.005$, $0     $) & \multirow{4}{*}{BN + nonlinearity} \\
                   &                         &                          &                                              & ($1$, $0.600$, $-2...2$) \\
                   &                         &                          &                                              & ($2$, $0.600$, $-3...3$) \\
                   &                         &                          &                                              & ($3$, $0.400$, $-2...2$) \\
\addlinespace[0.5em]
\multirow{3}{*}{6} & \multirow{3}{*}{Conv2D} & \multirow{3}{*}{crop(2)} & \multirow{3}{*}{$ 1* 1*96*c_\textrm{final}$} & ($0$, $0.005$, $0     $) & \multirow{3}{*}{BN + nonlinearity} \\
                   &                         &                          &                                              & ($1$, $0.600$, $-2...2$) \\
                   &                         &                          &                                              & ($2$, $0.400$, $-2...2$) \\
\midrule

		            7  &                 Linear  & --                       &                 $      96                 $  & -- & BN + nonlinearity \\
		            8  &                 Linear  & --                       &                 $      96                 $  & -- & BN + nonlinearity \\
		            9  &                 Linear  & --                       &                 $      40                 $  & -- & CrossEntropyLoss  \\
		\bottomrule
		\label{tab:detailed_mnist_arch}
  \end{tabular}
\end{table}

\textbf{Experiments:}
We train different variations of our model on {\mnistrot} (Table~\ref{tab:mnist_arch_variations}). We measure accuracy and training time per epoch with varying precision of the angular representation by using a different number of Fourier coefficients (3, 5, 9, or 17), as well as precision of the nonlinearity application by using different padding sizes for the FFT algorithm. 

As \emph{invariance operation}, we consider two possible choices. One method to produce invariant activations, \emph{conv2triv}, is to output only the rotation invariant coefficient $z_0$ in the last convolutional layer (i.e. to set $c_\textrm{final} = 1$ in Table~\ref{tab:mnist_arch_variations}). This coefficient is the average value of the angular function, therefore this method is equivalent to average pooling over different rotations of the input. However, as strong activations may appear only for some specific rotation of the input, this method may not be optimal, as can be also seen from the results. As an alternative, we set the last convolutional layer to full sized output ($c_\textrm{final} = c_\textrm{max}$) and take the \emph{norm} of each complex coefficient after the nonlinearity has been applied.

\begin{table}
  \caption{Architectural variations of our {\mnistrot} model}
  \centering
	\scriptsize
  \begin{tabular}{ccccccccccc}
    \toprule
    \multirow{2}{*}{model} &
		\multirow{2}{*}{group} & repre-    & num.   & FFT & activation            & invariant        & model     & sec / & \multicolumn{2}{c}{test error (\%)} \\
          &                & sentation & coeff. & pad & function              & map              & param.    & epoch & mean  & std   \\
		\midrule
		Ours  & $SO(2)$        & Fourier   &  3     & 127 & $\ReLU$               & \emph{norm}      &   365,338 & 21   & 0.985 & 0.035 \\
		\midrule
		Ours  & $SO(2)$        & Fourier   &  5     & 127 & $\ReLU$               & \emph{norm}      &   708,634 & 25   & 0.768 & 0.021 \\
		\midrule
		Ours  & $SO(2)$        & Fourier   &  9     &   0 & $\ReLU$               & \emph{norm}      & 1,394,986 & 30   & 0.710 & 0.025 \\
		Ours  & $SO(2)$        & Fourier   &  9     &   7 & $\ReLU$               & \emph{norm}      & 1,394,986 & 30   & 0.689 & 0.019 \\
		Ours  & $SO(2)$        & Fourier   &  9     & 127 & $\ReLU$               & \emph{norm}      & 1,394,986 & 36   & 0.685 & 0.026 \\
		Ours  & $SO(2)$        & Fourier   &  9     & 127 & $\ReLU$               & \emph{conv2triv} &   891,178 & 36   & 0.719 & 0.018 \\
		\midrule
    Ours  & $SO(2)$        & Fourier   & 17     & 127 & $\ReLU$               & \emph{norm}      & 2,729,098 & 64   & 0.699 & 0.033 \\
		\bottomrule
		\label{tab:mnist_arch_variations}
  \end{tabular}
\end{table}

Having found a well performing architecture, we use it to investigate different nonlinearities (Table~\ref{tab:mnist_nonlinearity_variations}). We evaluate the $\ReLU$, Leaky$\ReLU$, $\SiLU$, $\ELU$, $\tanh$ and $\Sigmoid$ nonlinearities with the FFT algorithm using a padding of $127$ to guarantee a good precision, with $\ReLU$ and Leaky$\ReLU$ resulting in the best accuracy. We also include the polynomial approximations of $\ReLU$~\cite{Gottemukkula2020PolynomialActications}, where we set the FFT padding depending on the degree of the polynomial. These approximations were designed for an input range of $[-5;5]$. As polynomials can reach very high values outside of the intended range, we clamp the L1-norm of each channel's Fourier coefficients (which is an upper limit to the maximum absolute value of the angular function) to a maximum value of 5 before applying the nonlinearity, to make sure we avoid problems with exploding activations or gradients.

Another method is to use norm-only nonlinearities, which do not require a Fourier transform as they act exclusively on the norm of the complex coefficients~\cite{worrall2017harmonic,Weiler-generalE2-NeurIPS19}. This is only suitable for functions with strictly positive output (like $\ReLU$ and $\Sigmoid$). In accordance with \citet{Weiler-generalE2-NeurIPS19}, we find that norm-based nonlinearities lead to a significantly worse accuracy.

\begin{table}
  \caption{Comparison of nonlinearities on {\mnistrot}}
  \centering
	\scriptsize
  \begin{tabular}{ccccccccccc}
    \toprule
    \multirow{2}{*}{model} &
		\multirow{2}{*}{group} & repre-    & num.   & FFT & activation         & invariant        & model     & sec / & \multicolumn{2}{c}{test error (\%)} \\
          &                & sentation & coeff. & pad & function           & map              & param.    & epoch & mean  & std   \\
		\midrule
		Ours  & $SO(2)$        & Fourier   &  9     &  -- & $\doubleC$-$\ReLU$    & \emph{norm}      & 1,396,138 & 30   & 0.980 & 0.031 \\
		Ours  & $SO(2)$        & Fourier   &  9     &  -- & $\doubleC$-$\Sigmoid$ & \emph{norm}      & 1,396,138 & 30   & 1.500 & 0.034 \\
		\midrule
		Ours  & $SO(2)$        & Fourier   &  9     & 127 & $\ReLU$               & \emph{norm}      & 1,394,986 & 36   & 0.685 & 0.026 \\
		Ours  & $SO(2)$        & Fourier   &  9     & 127 & Leaky$\ReLU$          & \emph{norm}      & 1,394,986 & 36   & 0.690 & 0.028 \\
		Ours  & $SO(2)$        & Fourier   &  9     & 127 & $\SiLU$               & \emph{norm}      & 1,394,986 & 36   & 0.705 & 0.026 \\
		Ours  & $SO(2)$        & Fourier   &  9     & 127 & $\ELU$                & \emph{norm}      & 1,394,986 & 36   & 0.729 & 0.029 \\
		Ours  & $SO(2)$        & Fourier   &  9     & 127 & $\tanh$               & \emph{norm}      & 1,394,986 & 36   & 0.768 & 0.024 \\
		Ours  & $SO(2)$        & Fourier   &  9     & 127 & $\Sigmoid$            & \emph{norm}      & 1,394,986 & 36   & 0.809 & 0.022 \\
		\midrule
		Ours  & $SO(2)$        & Fourier   &  9     &   8 & Poly(2)               & \emph{norm}      & 1,394,986 & 32   & 0.690 & 0.015 \\
		Ours  & $SO(2)$        & Fourier   &  9     &  24 & Poly(4)               & \emph{norm}      & 1,394,986 & 36   & 0.690 & 0.024 \\
    \bottomrule
		\label{tab:mnist_nonlinearity_variations}
  \end{tabular}
\end{table}

\begin{figure}
  \centering
	\includegraphics[width=\linewidth]{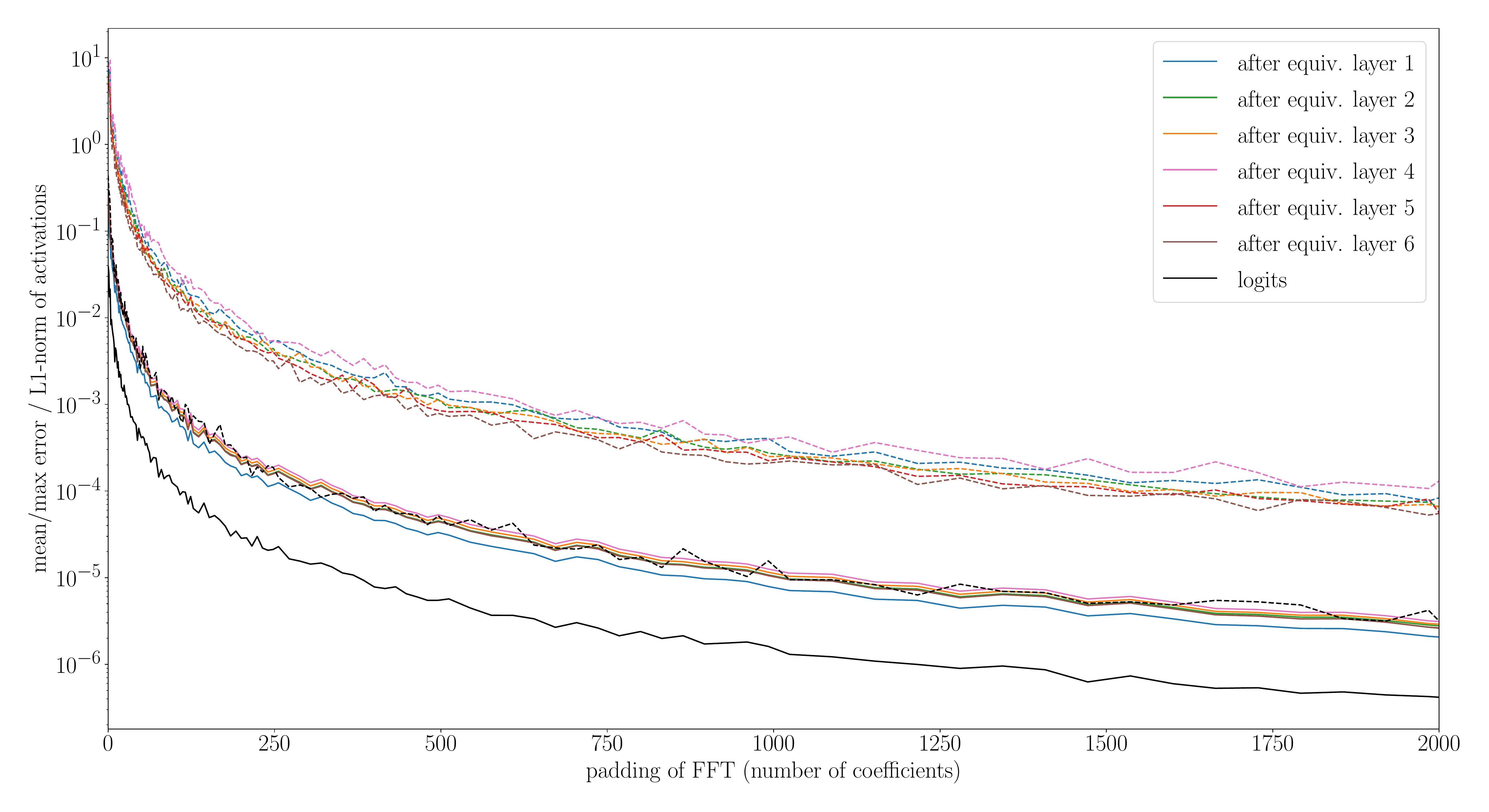}
  \caption{Relative error of $\ReLU$ activations for our 2D architecture measured on {\mnistrot} for randomly rotated vs. unrotated inputs (architecture as in Table~\ref{tab:detailed_mnist_arch}, 9 Fourier coefficients, \emph{norm}-map). solid: mean absolute error, dashed: maximum error, relative to the layer-wise L1-norm measured for batches of 32 images, with 36 random rotations applied to each image}
	\label{fig:errorReLUlarge}
\end{figure}

\begin{figure}
  \centering
	\includegraphics[width=\linewidth]{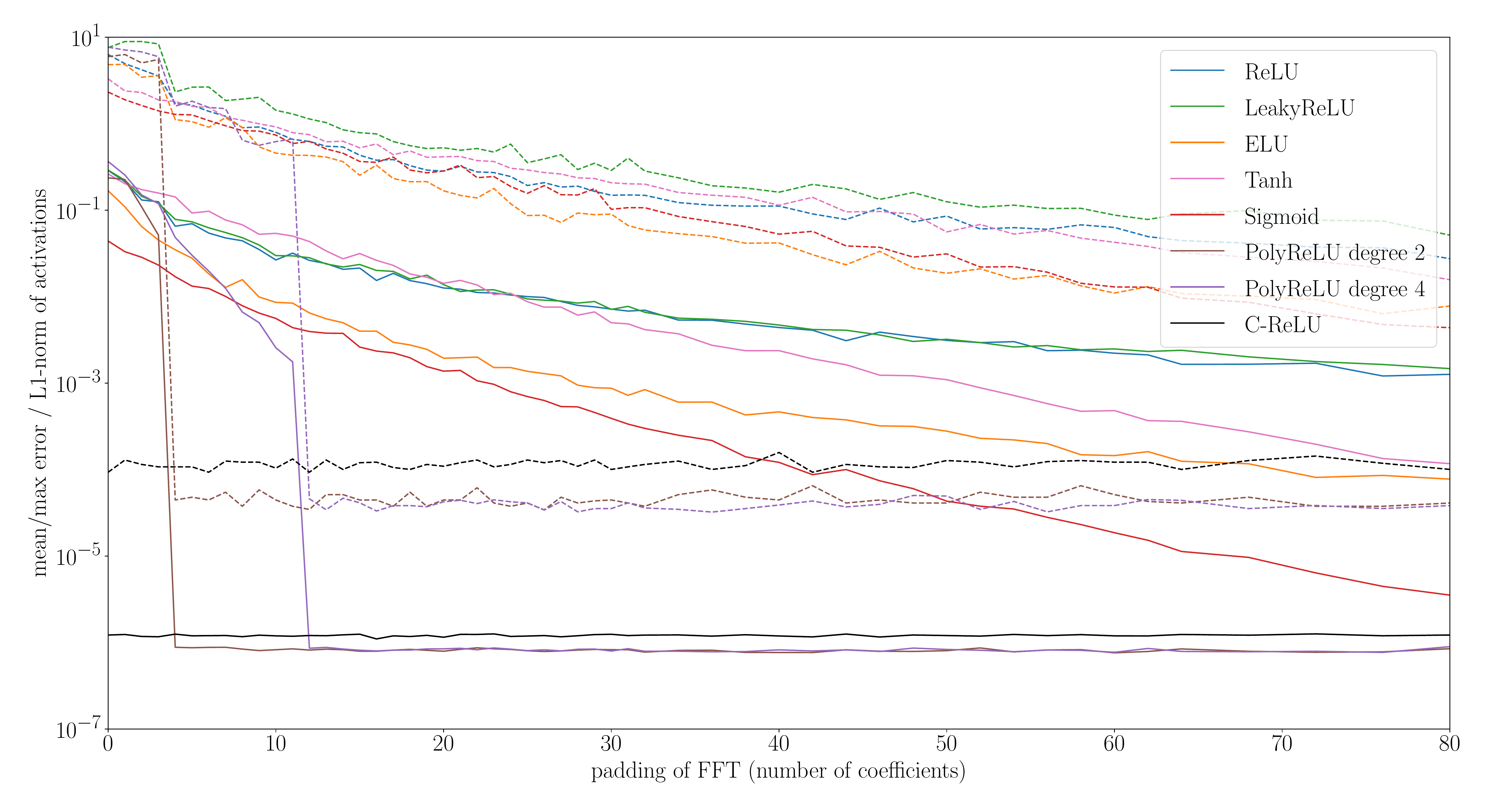}
  \caption{Relative error for various nonlinearities for our 2D architecture measured on {\mnistrot} for randomly rotated vs. unrotated inputs. Errors are measured as in Fig.~\ref{fig:errorReLUlarge} after the fifth (penultimate) equivariant layer. Polynomials show the expected sharp decline at with increasing FFT padding. $\doubleC$-$\ReLU$ included as reference (no FFT used).}
	\label{fig:errorVarious80large}
\end{figure}

\begin{figure}
  \centering
	\includegraphics[width=\linewidth]{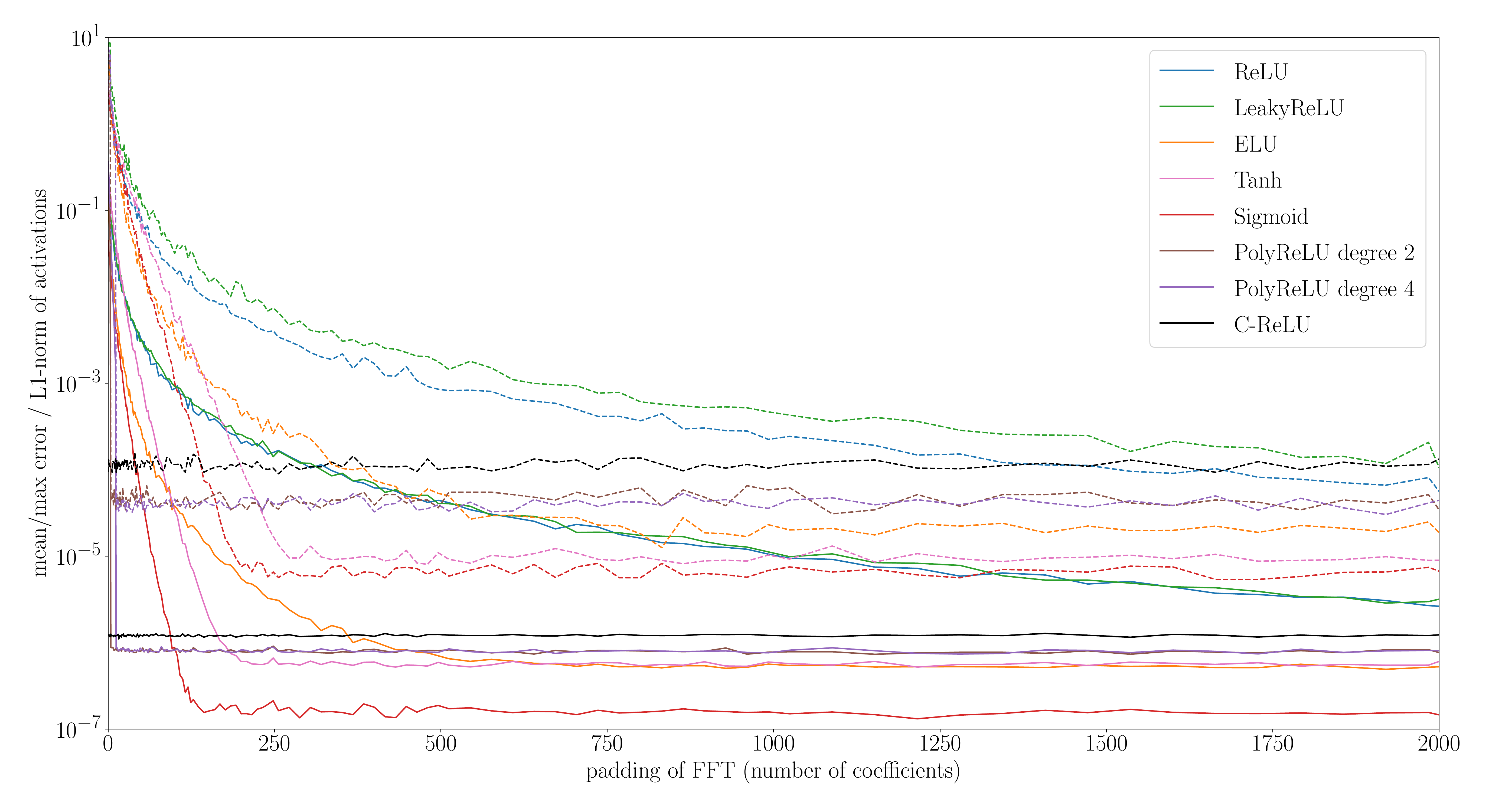}
  \caption{Same plot as Figure~\ref{fig:errorVarious80large}, but extended to very large FTT paddings. $\ReLU$ and Leaky$\ReLU$ converge slowest and still show improvements up to very large padding values of 2000.}
	\label{fig:errorVarious2000large}
\end{figure}

\subsection{$SE(3)$-Equivariant Surfel Networks}

Our $SE(3)$ based surfel network is inspired by the work of \citet{Wiersma-SurfaceCNNs-Siggraph-2020} and transfers the concept of angular-dependent activations, which previously lived on a flat surface of an image, to the curved surface of a 3D object. This poses some difficulties, as the implicit assumption that all features are aligned in an identical way (which we made for the 2D case of flat images) does not hold for generally curved 3D surfaces, where the normal vectors can have varying directions. Therefore, this approach requires to assign a \emph{local reference coordinate frame} ($\textbf{x}$, $\textbf{y}$, $\textbf{n}$) to each point feature, consisting of the normal vector $\textbf{n}$, an arbitrarily chosen tangential vector $\textbf{x}\perp\textbf{n}$, and a second tangential vector $\textbf{y} = \textbf{n}\times\textbf{x}$ (calculated by taking across product, to obtain a right-handed coordinate system). The angular features $x(\alpha)$ can then be mapped to directions $\textbf{v}$ in the tangential $\textbf{x}$,$\textbf{y}$-plane:
\begin{equation}
	\textbf{v} = \textbf{x}\cos\alpha - \textbf{y}\sin\alpha
	\label{eq:direction}
\end{equation}

For two features to interact in an equivariant fashion, we need to take their local alignment into account. \citet{Wiersma-SurfaceCNNs-Siggraph-2020} solve this problem by using parallel transport along the surface of the object to align the two local coordinate systems in a reliable manner. We chose to take a different approach. We first find a common tangent vector $\textbf{u}$ which lies in the tangential $\textbf{x}$,$\textbf{y}$-plane of both (input and output) local coordinate systems and rotate all input Fourier features to this common coordinate system. We then multiply the imaginary part of all coefficients by the dot product of the normal vectors $\langle \textbf{n}_1, \textbf{n}_2 \rangle$, after which we rotate the coefficients to the coordinate system of the output. For coefficients of rotation order 1 ($z_1$), where activations can be represented as tangential vectors fixed to the object that point in the direction of maximal activation according to Equation~\ref{eq:direction}, this approach is indentical to projecting these vectors to the tangential plane of the output point, as depicted in Figure~3 in the main paper.%~\ref{fig:proj-tangent}.

\begin{table}
  \caption{Detailed architecture for experiments on the {\modelnetforty} dataset}
  \centering
	\scriptsize
  \begin{tabular}{cccccccccc}
\toprule
\multirow{2}{*}{no.} & \multirow{2}{*}{layer type} & \multicolumn{3}{c}{layer output} & \multicolumn{3}{c}{layer filters}& \multirow{2}{*}{followed by} \\ \cmidrule(r){3-5} \cmidrule(r){6-8}
								&            & channels & coeffs &sampling level& radius& levels                 &$\sigma$&                   \\
\midrule
0               & Input      & 1        & 1      & 1            & --    & --                     & --     & --                \\
1               & SurfelConv & 16       & 9      & 2            & $0.1$ & ($-0.1$, $0.1$, $0.1$) & 0.0424 & BN + nonlinearity \\
2               & SurfelConv & 32       & 9      & 3            & $0.2$ & ($-0.2$, $0.2$, $0.2$) & 0.0849 & BN + nonlinearity \\
3               & SurfelConv & 48       & 9      & 3            & $0.4$ & ($-0.4$, $0.4$, $0.4$) & 0.1699 & BN + nonlinearity \\
4               & SurfelConv & 64       & 9      & 3            & $0.4$ & ($-0.4$, $0.4$, $0.4$) & 0.1699 & BN + nonlinearity \\
5               & SurfelConv & 96       & 9      & 3            & $0.8$ & ($-0.8$, $0.8$, $0.8$) & 0.3397 & BN + nonlinearity \\
6               & SurfelConv & 40       & 1      & 3            & $0.8$ & ($-0.8$, $0.8$, $0.8$) & 0.3397 & BN + nonlinearity \\
7               & PointAvgPool & 40     & 1      & (1 value)    & --    & --                     & --     & CrossEntropyLoss  \\
\bottomrule
		\label{tab:detailed_modelnet_arch}
  \end{tabular}
\end{table}

\textbf{Data set:} For the 3D surfel case, we use {\modelnetfortylink} as benchmark. We rescale all models to fit in a bounding cube (coordinate range $[-1,1]$) while keeping the aspect ratio and convert the polygonal data into point clouds by z-Buffer rasterization with high resolution from 50 random view points. This method keeps only points on the surface of the model which are visible from the outside, which we found to slightly improve performance, as some models in {\modelnetforty} contain a lot of internal elements. Normals are estimated from a PCA-fit to 20 nearest neighbors at a sample spacing of $0.005$, with some small random dithering value added to avoid numerical instabilities in case when two normals have the exact same direction, and always oriented to point away from origin of the coordinate system. 

The final input to our network is obtained by a by Poisson disc sampling with a sample spacing of $0.05$ for sampling level 1. Subsequent sampling levels are created by iteratively, choosing a quarter of the points of the previous sampling level in each step. For this process, we use the sample elimination algorithm described by \citet{Yuksel2015}.
Note that the final Poisson disc sampling process is done individually for each training epoch, which each time generates different samplings and thus is important to avoid overfitting on a specific sampling in the later training process. As we use external tools for the preprocessing of the dataset, these steps are not included in our published code. The preprocessed data to repeat our experiments will be provided on request.

As our preprocessing only gives us a set of input points and corresponding normals, but not any initial feature vectors associated with these points, we use a single-channels input tensor with the rotation-invariant Fourier coefficient set to 1 for each point as input.

\textbf{Architecture:}
Our detailed architecture is shown in Table~\ref{tab:detailed_modelnet_arch}. Experiments have shown that convolutional layers with a filter stack of 3 equidistant levels (Fig.~2 in the main paper), %~\ref{fig:surfel-network},
where each level consists of a single filter ring with a radius equal to the spacing between adjacent levels, work well in practice. The $\textrm{FWHM}$ (full width a half maximum) of the Gaussian filter profile is also chosen equal to to the level spacing, resulting in the $\sigma$ values given in Table~\ref{tab:detailed_modelnet_arch} ($\sigma = \textrm{FWHM} / \sqrt{8\ln 2}$). We use 9 coefficients to represent our angular dependent functions and we do not impose limits on filter frequencies, i.e. allow interactions between all coefficients, which results in filter frequencies of a rotation order of up to 8. Our network consists of 6 equivariant convolutional layers. To generate rotation invariant output, we output only the invariant coefficients in the final convolutional layer (\emph{conv2triv}) and apply average pooling over all remaining points. As in the {\mnistrot} scenario, we use Batch Normalization (directly on Fourier coefficients, if applicable) and a nonlinearity after each convolutional layer except the last layer.

\begin{figure}
  \centering
	\includegraphics[width=\linewidth]{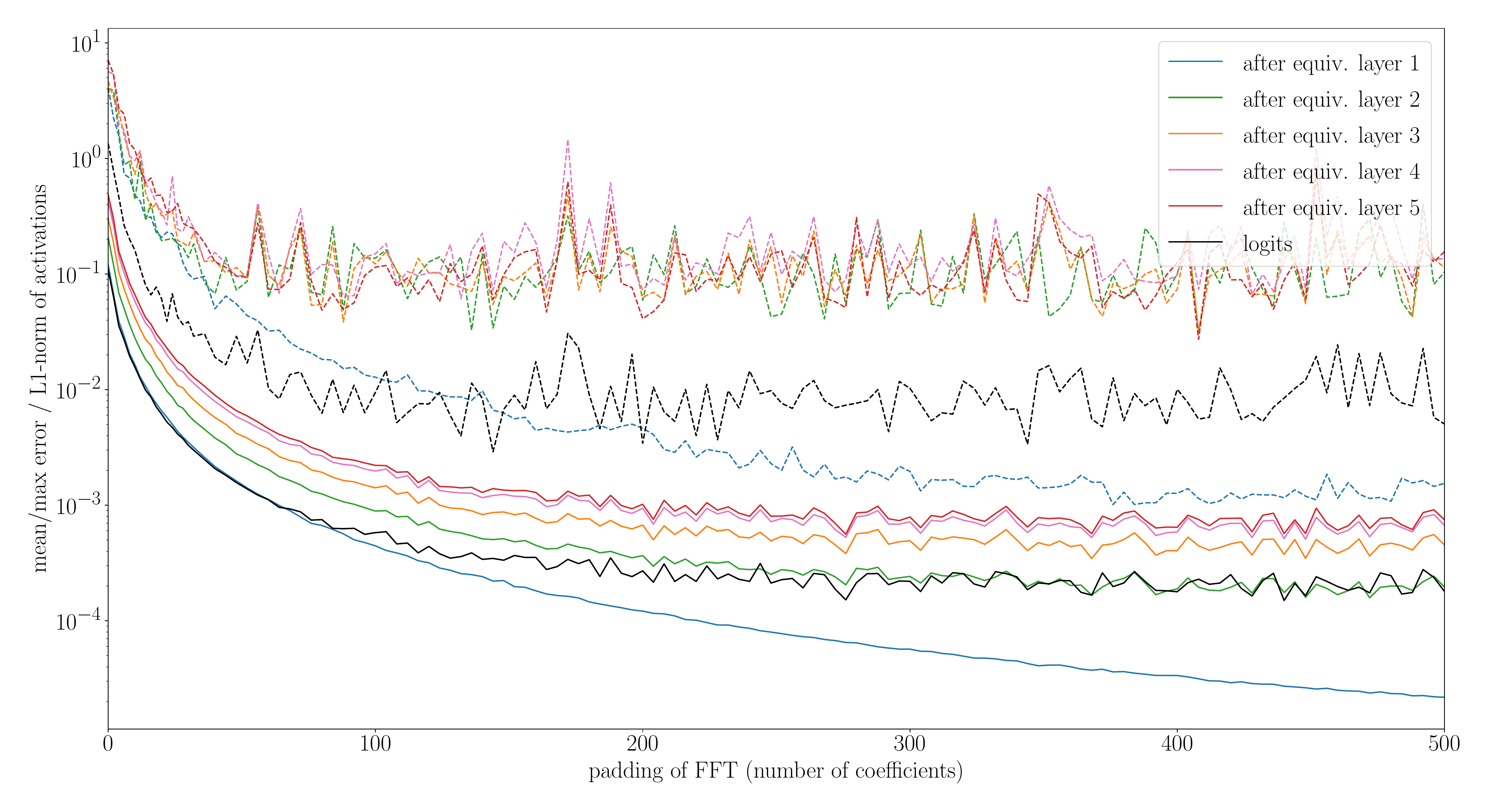}
  \caption{Relative error of $\ReLU$ activations for our 3D surfel architecture measured on {\modelnetforty} for randomly rotated vs. unrotated inputs (architecture as in Table~\ref{tab:detailed_modelnet_arch}). Note the different scale than the on the axes compared to the {\mnistrot} plots. solid: mean absolute error, dashed: maximum error, relative to the layer-wise L1-norm measured for batches of 32 images where a random $SO(3)$ rotation is applied individually for each image}
	\label{fig:errorReLUlargeModelnet}
\end{figure}

\begin{figure}
  \centering
	\includegraphics[width=\linewidth]{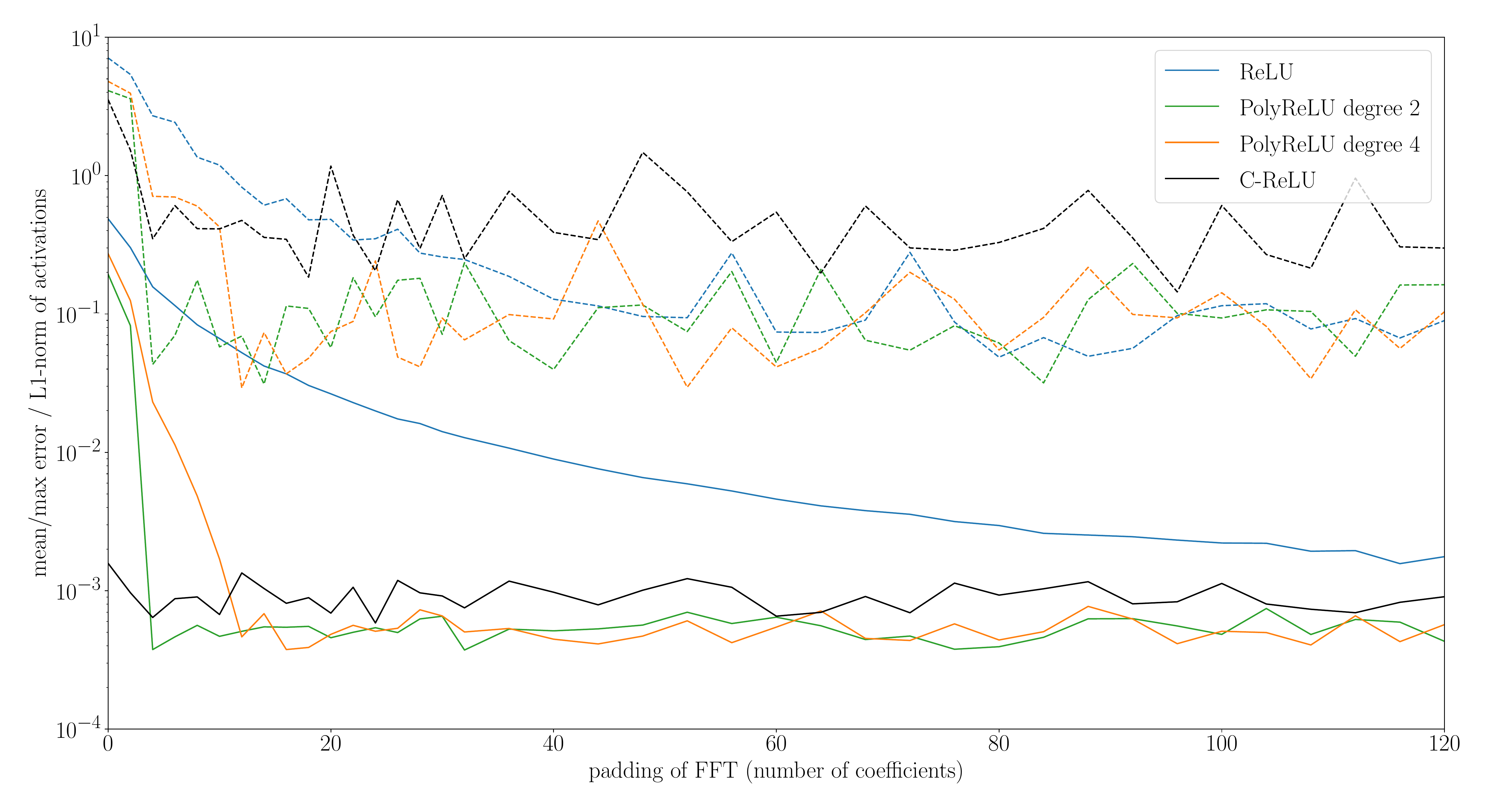}
  \caption{Relative error for various nonlinearities for our 3D surfel architecture measured on {\modelnetforty} for randomly rotated vs. unrotated inputs. Errors are measured as in Fig.~\ref{fig:errorReLUlargeModelnet} after the fifth (penultimate) convolutional layer. $\doubleC$-$\ReLU$ included as reference (no FFT used).}
	\label{fig:errorVariousLargeModelnet}
\end{figure}

\textbf{Training and testing:}
We train out networks on the training set of {\modelnetforty}, containing 9,843 3D meshes of 40 different object classes. There is a significant imbalance of the number of objects per class between training and testing set, however, we chose not to apply any measures to counteract this, to allow a fair comparison of our results to those of other papers.

We initialize the network weights randomly before training using He initialization~\cite{HeInit}, and train using the Adam optimizer with \emph{PyTorch} default parameters ($\textrm{betas}=(0.9, 0.999)$, $\textrm{eps}=10^{-8}$) and a mini-batch size of 32 to optimize the log-softmax cross entropy of the network output. Training is done for a total of 30 epochs, using a fixed learning rate of 0.015 in the first 10 epochs, an then decayed linearly to zero over the next 20 epochs. Batch statistics are calculated over the full training set, as outlined in Section~\ref{sec:appendix_mnist}.

We test our architecture on the testing set of {\modelnetforty}, containing 2,468 3D models. As common in the literature, we perform runs with different rotational augmentation during training and testing,
\begin{itemize}
	\item N/$SO(3)$, referring to training on the original (non-augmented) dataset and testing with random $SO(3)$ rotations,
	\item z/$SO(3)$, denoting random rotations around the $z$-axis during training and $SO(3)$ rotations during testing, and
	\item $SO(3)$/$SO(3)$, using random $SO(3)$ augmentation during training and testing.
\end{itemize}
Our results for {\modelnetforty}, including runtimes (measured without taking the preprocessing steps into account), can be found in Table~2 in the main paper. %~\ref{tab:modelnet40}.
Rotation error measurements for random rotations can be found in Figure~\ref{fig:errorReLUlargeModelnet} for $\ReLU$ and Figure~\ref{fig:errorVariousLargeModelnet} for other nonlinearities. The error of is generally higher than for our 2D architecture, as can be seen from the layerwise plot in Figure~\ref{fig:errorReLUlargeModelnet}. It increases in deeper layers, probably due to cumulative effects. However, the network is still mostly invariant in the classification task, as can be seen from Table~2 in the main paper. %~\ref{tab:modelnet40}.

The higher error is most likely not caused by the FFT algorithm for applying nonlinearities, as $\doubleC$-$\ReLU$ (which is equivariant by construction, and thus does not use FFT calculations at all) produces a similar error level (see Figure~\ref{fig:errorVariousLargeModelnet}). Therefore, we assume this is a general drawback of the 3D surfel architecture used in this test, probably due to the much more complex calculations in comparison to the 2D case and thus higher amplification of noise. As the relative error is the same (polynomials) or easily approaches the same level (ReLU) as the fully-equivariant baseline ($\doubleC$-$\ReLU$), this does not directly weaken the result of our paper (that more general nonlinearities can be handled at error rates comparable to specialized solutions with provable equivariance without additional measures). However, it indicates that other sources of numerical error amplification might be a separate issue to pay attention to if strong invariances need to be maintained.

\clearpage

%\section*{Acknowledgements}
%This work was supported by the Collaborative Research Center TRR 146 of the Deutsche Forschungsgemeinschaft (DFG).

{
\small
\bibliography{nonlinearities-s2021}
}